\documentclass{article}

\usepackage{microtype}
\usepackage{graphicx}
\usepackage{subfigure}
\usepackage{booktabs} % for professional tables

\usepackage{hyperref}

\usepackage{amsfonts}
\usepackage{amsmath}
\usepackage{empheq}
\usepackage{mathtools}
\usepackage{xspace}
\usepackage{amsthm}
\usepackage{enumitem}
\setlist{nolistsep}

\newcommand{\eg}{\textit{e.g.}\xspace}
\newcommand{\ie}{\textit{i.e.}\xspace}
\newcommand{\iid}{\textit{i.i.d.}\xspace}

\newtheorem{theorem}{Theorem}

\newtheorem{lemma}{Lemma}

\usepackage[accepted]{icml2018}

\icmltitlerunning{Error Compensated Quantized SGD and its Applications to Large-scale Distributed Optimization}

\begin{document}

\twocolumn[
\icmltitle{Error Compensated Quantized SGD and its Applications to Large-scale Distributed Optimization}

% Optimized Quantized Stochastic Gradient Descent for Efficient Distributed Optimization



\icmlsetsymbol{equal}{*}

\begin{icmlauthorlist}
\icmlauthor{Jiaxiang Wu}{tencent}
\icmlauthor{Weidong Huang}{tencent}
\icmlauthor{Junzhou Huang}{tencent}
\icmlauthor{Tong Zhang}{tencent}
\end{icmlauthorlist}

\icmlaffiliation{tencent}{Tencent AI Lab, Shenzhen, China}

\icmlcorrespondingauthor{Jiaxiang Wu}{jonathanwu@tencent.com}

\icmlkeywords{Distributed Optimization, Stochastic Gradient Descent, Gradient Quantization}

\vskip 0.3in
]



\printAffiliationsAndNotice{}  % leave blank if no need to mention equal contribution

\begin{abstract}
Large-scale distributed optimization is of great importance in various applications. For data-parallel based distributed learning, the inter-node gradient communication often becomes the performance bottleneck. In this paper, we propose the error compensated quantized stochastic gradient descent algorithm to improve the training efficiency. Local gradients are quantized to reduce the communication overhead, and accumulated quantization error is utilized to speed up the convergence. Furthermore, we present theoretical analysis on the convergence behaviour, and demonstrate its advantage over competitors. Extensive experiments indicate that our algorithm can compress gradients by a factor of up to two magnitudes without performance degradation.
\end{abstract}

\section{Introduction}

Due to the explosive growth of data in recent years, large-scale machine learning has attracted increasing attention in various domains, such as computer vision and speech recognition. Distributed optimization is one of the core building blocks in these applications, where the training data is often too massive to be efficiently handled by a single computation node.

A commonly used distributed learning framework is data parallelism, in which the whole data set is split and stored on multiple nodes within a cluster. Each node computes its local gradients and communicates gradients with other nodes to update model parameters. For such learning systems, the time consumption can be roughly categorized as either computation or communication. The communication often becomes the performance bottleneck, especially for large clusters and/or models with tons of parameters.

There have been several works attempting to improve the efficiency of distributed learning by reducing the communication cost. Some methods focused on quantizing gradients into the fixed-point numbers \cite{zhou2016dorefa, alistarh2017qsgd}, so that much fewer bits are needed to be transimitted. More aggressive quantization, such as the binary or ternary representation, has also been investigated in \cite{seide2014bit, strom2015calable, wen2017terngrad}. Other methods imposed sparsity onto gradients during communication, where only a small fraction of gradients get exchanged across nodes in each iteration \cite{wangni2017gradient, lin2018deep}.

The underlying ideas of these methods are basically to compress gradients into some special form, in which each entry can be represented by much fewer bits than the original 32-bit floating-point number. Such compression introduces extra stochastic noises, \ie quantization error, into the optimization process, and will slow down the convergence or even leads to divergence. 1Bit-SGD \cite{seide2014bit} adopted the error feedback scheme, which was to compensate the current local gradients with quantization error from the last iteration, before feeding it into the quantization function. Although authors stated that this improved the convergence behaviour, no theoretical analysis was given to evidence its effectiveness.

In this paper, we propose the error compensated quantized stochastic gradient descent method, namely ECQ-SGD. Our algorithm also utilizes the error feedback scheme, but here we accumulate all the previous quantization errors, rather than only from the last iteration as in 1Bit-SGD. Although empirical evaluation shows that this modification leads to faster and more stable convergence than many baseline methods, it is non-trivial to establish the theoretical guarantee for this phenomenon.

In \cite{alistarh2017qsgd}, authors proved that for their proposed QSGD algorithm, the number of iterations required to reach a certain sub-optimality gap is proportional to the variance bound of stochastic quantized gradients. However, this cannot explain our method's convergence behaviour, since our quantized gradients are biased estimations, unlike in QSGD. Actually, the variance bound of quantized gradients is even larger than that in QSGD, due to the accumulated quantization error. In order to address this issue, we present the convergence analysis from another perspective, and prove that our algorithm has a tighter worst-case error bound than QSGD with properly chosen hyper-parameters. It can be shown that our proposed error feedback scheme can well suppress the quantization error's contribution to the error bound, leading to a smaller sub-optimality gap than QSGD, as we observed in experiments.

The remaining part of this paper will be organized as follows. We briefly review related works in Section 2, and present preliminaries in Section 3. In Section 4 and 5, we propose the ECQ-SGD algorithm and its theoretical analysis. Extensive experiments are carried out in Section 6, followed by the final conclusions.

\section{Related Works}

There have been several works proposed to speed up stochastic gradient descent in the context of distributed learning. Some adopt asynchronous update to decouple computation from communication, while others focus on reducing the communication overhead with gradient quantization or sparsification.

\textbf{Asynchronous SGD.} Hogwild! \cite{recht2011hogwild} is a lock-free implementation of paralleled SGD that can achieve a nearly optimal rate of convergence for certain problems. \cite{dean2012large} proposes the \textit{DistBelief} framework, which adopts asynchronous SGD to train deep networks under a distributed setting. The convergence behaviour of asynchronous SGD have been extensively analysed in many works \cite{chaturapruek2015asynchronous, zhao2016fast, zheng2017asynchronous, desa2017understanding}.

\textbf{Gradient Quantization.} In \cite{seide2014bit}, 1Bit-SGD is proposed to quantize each gradient component to either 1 or -1 with zero-thresholding. An error feedback scheme is introduced during quantization, to compensate the quantization error from the last iteration. Similar ideas are adopted in \cite{strom2015calable}, which accumulates local gradients across iterations, and only transmits gradient components exceeding a pre-selected threshold. Wen et al. further extend this idea and compress gradients into ternary values with a stochastic quantization function to ensure the unbiasness \cite{wen2017terngrad}. Quantized SGD \cite{alistarh2017qsgd} randomly quantizes gradients using uniformly distributed quantization points, and detailed analysis is presented to address its convergence. ZipML \cite{zhang2017zipml} introduces an optimal quantization strategy via dynamically choosing quantization points based on the distribution. Zhou et al. propose the DoReFa-Net to train convolutional networks with inputs, weights, and gradients all quantized into fixed-point numbers \cite{zhou2016dorefa}.

\textbf{Gradient Sparsification.} The gradient dropping method is proposed in \cite{aji2017sparse} to introduce sparsity into gradients to reduce the communication cost. In \cite{wangni2017gradient}, gradient sparsification is modelled as a linear programming problem, aiming to minimize the variance increase of quantized gradients. Lin et al. propose the deep gradient compression algorithm, utilizing momentum correction, gradient clipping, momentum factor masking, and warm-up training to achieve higher sparsity without losing the accuracy \cite{lin2018deep}.

\section{Preliminaries}

We consider the following unconstrained optimization:
\begin{equation}
\min_{\mathbf{w}} ~ f \left( \mathbf{w} \right)
\end{equation}
where $\mathbf{w} \in \mathbb{R}^{d}$ and $f: \mathbb{R}^{d} \rightarrow \mathbb{R}$ is a convex and differentiable function we wish to minimize. Often, the objective function $f$ is defined on a set of training samples $\mathcal{D} = \left\{ \mathbf{x}_{i} \right\}$, and the need for distributed optimization arises when the training set is too large to fit into a single node.

Assume we are solving this distributed optimization problem in a data-parallel manner. The full set $\mathcal{D}$ is evenly distributed across $P$ nodes, and the data subset located at the $p$-th node is denoted as $\mathcal{D}_{p}$. Formally, we wish to optimize:
\begin{equation}
\min_{\mathbf{w}} ~ \sum_{p = 1}^{P} \sum_{\mathbf{x}_{i} \in \mathcal{D}_{p}} f \left( \mathbf{w}; \mathbf{x}_{i} \right)
\end{equation}

\begin{figure}[ht]
\begin{center}
\centerline{\includegraphics[width=.8\columnwidth]{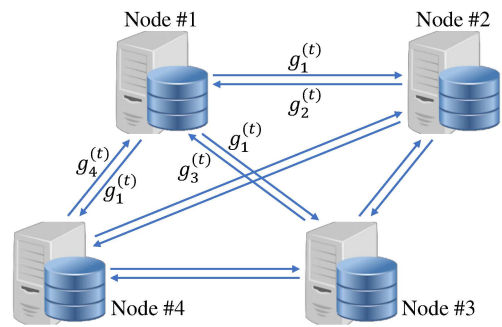}}
\caption{Distributed optimization under the data-parallel setting.}
\label{fig:distributed_optimization}
\end{center}
\end{figure}

Figure \ref{fig:distributed_optimization} depicts how model parameters $\mathbf{w}$ are updated via distributed SGD. Every node initializes its local model replica using the same random seed, to ensure the consistency of all model replicas. In the $t$-th iteration, each node randomly selects a mini-batch of training samples, computes local gradients, and then broadcasts to all the other nodes. When one node gathers all the local gradients sent by other nodes, global gradients can be computed and used to update model parameters.

\section{Error Compensated Quantized SGD}

For distributed optimization under the data-parallel setting, it is required to exchange local gradients between every two nodes in each iteration. For large-scale distributed optimization with massive number of model parameters, \eg training a convolutional neural network, gradient communication may become the performance bottleneck.

One possible solution is to quantize local gradients before transmission, so as to reduce the communication cost. In this section, we propose the ECQ-SGD (Error Compensated Quantized SGD) algorithm, which updates model parameters with quantized local gradients. In each iteration, current local gradients are compensated with accumulated quantization error from all the previous iterations, and then fed into a stochastic quantization function for compression.

Let $Q: \mathbb{R}^{d} \rightarrow \mathcal{C}^{d}$ be an unbiased stochastic quantization function, which maps each component in a $d$-dimensional vector into some element from the quantization codebook $\mathcal{C}$. The codebook usually only contains limited number of elements, so the quantized vector can be efficiently encoded. In each iteration, each node quantizes its local gradients before broadcasting:
\begin{equation}
\tilde{\mathbf{g}}_{p}^{( t )} = Q ( \mathbf{g}_{p}^{( t )} )
\end{equation}
where $\mathbf{g}_{p}^{( t )}$ is the local gradients of the $p$-th node at the $t$-th iteration, and $\tilde{\mathbf{g}}_{p}^{( t )}$ is its quantized counterpart.

When a node receives all the local gradients sent by other nodes, it computes the global gradients and updates its local model replica via:
\begin{equation}
\mathbf{w}^{( t + 1 )} = \mathbf{w}^{( t )} - \eta \cdot \tilde{\mathbf{g}}^{( t )} = \mathbf{w}^{( t )} - \frac{\eta}{P} \sum_{p = 1}^{P} \tilde{\mathbf{g}}_{p}^{( t )}
\end{equation}
where $\eta > 0$ is the learning rate.

The core idea of ECQ-SGD is that when quantizing local gradients, both the current gradients and previously accumulated quantization error should be taken into consideration. Specifically, we use $\mathbf{h}_{p}^{( t )}$ to denote the accumulated quantization error of the $p$-th node at the $t$-th iteration:
\begin{equation}
\mathbf{h}_{p}^{( t )} = \sum_{t' = 0}^{t - 1} \beta^{t - 1 - t'} ( \mathbf{g}_{p}^{( t' )} - \tilde{\mathbf{g}}_{p}^{( t' )} )
\label{eqn:def_accu_quan_err}
\end{equation}
where $\beta$ is the time decaying factor ($0 \le \beta \le 1$). Note that the accumulated quantization error can be incrementally updated:
\begin{equation}
\mathbf{h}_{p}^{( t )} = \beta \mathbf{h}_{p}^{( t - 1 )} + ( \mathbf{g}_{p}^{( t - 1 )} - \tilde{\mathbf{g}}_{p}^{( t - 1 )} )
\label{eqn:updt_accu_quan_err}
\end{equation}
where $\mathbf{h}_{p}^{( 0 )} = \mathbf{0}$. The quantized local gradients are now computed by applying the quantization function to the compensated gradients:
\begin{equation}
\tilde{\mathbf{g}}_{p}^{( t )} = Q ( \mathbf{g}_{p}^{( t )} + \alpha \mathbf{h}_{p}^{( t )} )
\label{eqn:quan_func_with_h}
\end{equation}
where $\alpha$ is the compensation coefficient ($\alpha \ge 0$).

Here, we adopt a stochastic quantization function with uniformly distributed quantization points, similar to QSGD \cite{alistarh2017qsgd}, where the $i$-th component is quantized as:
\begin{equation}
\tilde{g}_{i} = \left\| \mathbf{g} \right\| \cdot \text{sgn} \left( g_{i} \right) \cdot \xi \left( \left| g_{i} \right|; \left\| \mathbf{g} \right\| \right)
\end{equation}
where $\left\| \mathbf{g} \right\|$ acts as the scaling factor (possible choices include $l_{2}$-norm and $l_{\infty}$-norm), and $\xi \left( \cdot \right)$ is a stochastic function which maps a scalar to some element in $\left\{ 0, \frac{1}{s}, \dots, 1 \right\}$:
\begin{equation}
\xi \left( \left| g_{i} \right|; \left\| \mathbf{g} \right\| \right) = 
\begin{cases}
\frac{l}{s} ,&~ \text{with probability~} l + 1 - s \cdot \frac{\left| g_{i} \right|}{\left\| \mathbf{g} \right\|} \\
\frac{l + 1}{s} ,&~ \text{otherwise}
\end{cases}
\end{equation}
when $\left| g_{i} \right| / \left\| \mathbf{g} \right\|$ falls in the interval $\left[ \frac{l}{s}, \frac{l + 1}{s} \right)$. The hyper-parameter $s$ defines the number of non-zero quantization levels: a larger $s$ leads to more fine-grained quantization, at the cost of increased communication cost. From now on, we use $Q_{s} \left( \cdot \right)$ to denote the quantization function with $s$ non-zero quantization levels.

After quantization, we only need $r = \lceil \log_{2} \left( 2 s + 1 \right) \rceil$ bits to encode each quantized $\tilde{g}_{i}$, and one floating-point number to represent the scaling factor $\left\| \mathbf{g} \right\|$. The overall communication cost is $32 + d r$ bits ($r \ll 32$), which is much smaller than $32 d$ bits needed by the original 32-bit full-precision gradients. More efficient entropy encoding schemes, \eg Huffman encoding \cite{huffman1952method}, can further reduce the communication cost. Let $d_{k}$ denote the number of dimensions assigned to the $k$-th quantization level, then the overall encoding length is at most $\sum\nolimits_{k = 1}^{2 s + 1} d_{k} \log_{2} \frac{d}{d_{k}}$ bits.

\begin{algorithm}[tb]
\caption{Error Compensated Quantized SGD}
\label{alg:error_compensated_quantized_sgd}
\begin{algorithmic}
\STATE \textbf{Input:} distributed data $\mathcal{D} = \mathcal{D}_{1} \cup \cdots \cup \mathcal{D}_{P}$
\STATE initialize model parameters $\mathbf{w}^{( 0 )}$, and reset $\mathbf{h}_{p}^{( 0 )} \leftarrow \mathbf{0}$
\FOR{$t = 0, \dots, T - 1$}
\STATE // 1. Gradient Computation and Communication
\FOR{$p = 1, \dots, P$}
\STATE randomly select a mini-batch $\mathcal{D}_{p}^{( t )}$
\STATE compute local gradients $\mathbf{g}_{p}^{( t )} = \nabla f ( \mathbf{w}^{( t )}; \mathcal{D}_{p}^{( t )} )$
\STATE compute quantized local gradients $\tilde{\mathbf{g}}_{p}^{( t )}$ with (\ref{eqn:quan_func_with_h})
\STATE broadcast quantized local gradients $\tilde{\mathbf{g}}_{p}^{( t )}$
\STATE update the accumulated quantization error $\mathbf{h}_{p}^{( t + 1 )}$ with (\ref{eqn:updt_accu_quan_err})
\ENDFOR
\STATE
\STATE // 2. Model Update
\FOR{$p = 1, \dots, P$}
\STATE receive quantized local gradients $\{ \tilde{\mathbf{g}}_{p}^{( t )} \}$
\STATE compute global gradients $\tilde{\mathbf{g}}^{( t )} = \sum_{p} \tilde{\mathbf{g}}_{p}^{( t )} / P$
\STATE update model parameters $\mathbf{w}^{( t + 1 )} = \mathbf{w}^{( t )} - \eta \cdot \tilde{\mathbf{g}}^{( t )}$
\ENDFOR
\ENDFOR
\end{algorithmic}
\end{algorithm}

We summarize the overall workflow in Algorithm \ref{alg:error_compensated_quantized_sgd}. Since all the local gradients are quantized before transmission, the communication overhead can be greatly reduced. This is crucial to the learning efficiency, especially when the inter-node communication is the performance bottleneck.

\section{Theoretical Analysis}

In this section, we analyse the convergence behaviour of the proposed ECQ-SGD algorithm. We start with the discussion on the variance bound of quantization error. After that, we build up the error bound for quadratic optimization problems, and demonstrate ECQ-SGD's advantage over QSGD.

\subsection{Variance Bound of Quantization Error}

We define the quantization error as the difference between the compensated local gradients and its quantization results:
\begin{equation}
\boldsymbol{\varepsilon}_{p}^{( t )} = \tilde{\mathbf{g}}_{p}^{( t )} - \left( \mathbf{g}_{p}^{( t )} + \alpha \mathbf{h}_{p}^{( t )} \right)
\end{equation}

If the quantization function uses $l_{2}$-norm as the scaling factor, then it is identical to the one used in QSGD. In this case, we can directly borrow their conclusions on the following two properties of quantization error:

\begin{lemma}
\cite{alistarh2017qsgd} For any vector $\mathbf{v} \in \mathbb{R}^{d}$, let $\boldsymbol{\varepsilon} = Q_{s} \left( \mathbf{v} \right) - \mathbf{v}$ denote the quantization error, then we have:
\begin{itemize}
\itemsep 0pt
\item Unbiasness: $\mathbb{E} \left[ \boldsymbol{\varepsilon} \right] = \mathbf{0}$
\item Bounded variance: $\mathbb{E} \left\| \boldsymbol{\varepsilon} \right\|_{2}^{2} \le \min \left( \frac{d}{s^{2}}, \frac{\sqrt{d}}{s} \right) \cdot \left\| \mathbf{v} \right\|_{2}^{2}$
\end{itemize}
\end{lemma}

Here, we assume that the second moment of local gradients are bounded, \ie $\| \mathbf{g}_{p}^{( t )} \|_{2}^{2} \le B, \forall ( p, t )$. Under this assumption, the quantization error in QSGD satisfies:
\begin{equation}
\mathbb{E} \| \boldsymbol{\varepsilon}_{p}^{( t )} \|_{2}^{2} \le \gamma B
\end{equation}
where $\gamma = \min ( d / s^{2}, \sqrt{d} / s )$.

In ECQ-SGD, however, the vector to be quantized is a linear combination of current local gradients and accumulated quantization error. Still, we can derive a slightly relaxed bound for the quantization error's second moment:

\begin{lemma}
For the $p$-th node, its quantization error at the $t$-th iteration satisfies:
\begin{equation}
\mathbb{E} \| \boldsymbol{\varepsilon}_{p}^{( t )} \|_{2}^{2} \le \left( 1 + \alpha^{2} \gamma \cdot \frac{1 - \lambda^{t}}{1 - \lambda} \right) \cdot \gamma B
\end{equation}
where $\lambda = \alpha^{2} \gamma + ( \beta - \alpha )^{2}$.
\label{lem:opt_qsgd_quan_err_bnd}
\end{lemma}

\begin{proof}
Please refer to the supplementary material.
\end{proof}

In ECQ-SGD, we usually select hyper-parameters $( \alpha, \beta )$ to satisfy $\lambda < 1$ (\ie let $0 < \alpha < \frac{2}{\gamma + 1}$ and $\beta = 1$), ensuring that the following variance bound of quantization error holds for any iteration $t$:
\begin{equation}
\mathbb{E} \| \boldsymbol{\varepsilon}_{p}^{( t )} \|_{2}^{2} < \left( 1 + \frac{\alpha^{2} \gamma}{1 - \lambda} \right) \cdot \gamma B
\label{eqn:simple_variance_bound_comp}
\end{equation}

Therefore, as long as $\frac{\alpha^{2} \gamma}{1 - \lambda}$ is small, the variance bound of quantization error in the ECQ-SGD algorithm is still very close to that of QSGD. Also, it is worth noting that this upper bound does not depend on $t$, indicating that the variance bound will not diverge during the optimization.

Based on the above results, one can derive that the quantized local gradients' variance bound in ECQ-SGD is also larger than that in QSGD. According to the convergence analysis in QSGD, the number of iterations required to reach certain sub-optimality is proportional to the stochastic gradients' variance bound. This implies that following the convergence analysis in QSGD, one would conclude that ECQ-SGD should converge slower than QSGD, which actually conflicts with experimental results (as shown later). Therefore, we need to analyse ECQ-SGD's convergence behaviour from another perspective to explain its advantage over QSGD as observed in practice.

\subsection{Convergence for Quadratic Optimization}

Now we analyse how model parameters converge to the optimal solution in ECQ-SGD. Consider the following convex quadratic optimization:
\begin{equation}
\min_{\mathbf{w}} ~ \sum_{p = 1}^{P} \sum_{( \mathbf{A}_{i}, \mathbf{b}_{i} ) \in \mathcal{D}_{p}} \frac{1}{2} \mathbf{w}^{T} \mathbf{A}_{i} \mathbf{w} + \mathbf{b}_{i}^{T} \mathbf{w}
\end{equation}
where the whole data set $\mathcal{D} = \mathcal{D}_{1} \cup \dots \cup \mathcal{D}_{P}$ is evenly split and stored at $P$ nodes. By summing up all $\mathbf{A}_{i}$s and $\mathbf{b}_{i}$s, the above optimization is equivalent to:
\begin{equation}
\min_{\mathbf{w}} ~ \frac{1}{2} \mathbf{w}^{T} \mathbf{A} \mathbf{w} + \mathbf{b}^{T} \mathbf{w}
\end{equation}
where $\mathbf{A} = \sum_{( \mathbf{A}_{i}, \mathbf{b}_{i} ) \in \mathcal{D}} \mathbf{A}_{i}$ and $\mathbf{b} = \sum_{( \mathbf{A}_{i}, \mathbf{b}_{i} ) \in \mathcal{D}} \mathbf{b}_{i}$. The smallest and largest singular values of $\mathbf{A}$ are denoted as $a_{1}$ and $a_{2}$, respectively, \ie $a_{1} \mathbf{I} \preceq \mathbf{A} \preceq a_{2} \mathbf{I}$. Here, we assume $a_{1} > 0$ to ensure the strong convexity, which leads to the closed-form optimal solution $\mathbf{w}^{*} = - \mathbf{A}^{-1} \mathbf{b}$.

In each iteration, every node constructs a mini-batch of training samples $\{ ( \mathbf{A}_{i}, \mathbf{b}_{i} ) \}$, uniformly sampled from its local data subset. The resulting local gradients can be represented as the underlying true gradients plus a stochastic noise term, which is:
\begin{equation}
\mathbf{g}_{p}^{( t )} = \mathbf{A} \mathbf{w}^{( t )} + \mathbf{b} + \boldsymbol{\xi}_{p}^{( t )}
\end{equation}
where $\{ \boldsymbol{\xi}_{p}^{( t )} \}$ are \iid random noises with zero mean. The quantized local gradients are given by:
\begin{equation}
\tilde{\mathbf{g}}_{p}^{( t )} = \mathbf{A} \mathbf{w}^{( t )} + \mathbf{b} + \boldsymbol{\xi}_{p}^{( t )} + \alpha \mathbf{h}_{p}^{( t )} + \boldsymbol{\varepsilon}_{p}^{( t )}
\end{equation}

Recall the update rule of ECQ-SGD, we have:
\begin{equation}
\mathbf{w}^{( t + 1 )} = \mathbf{w}^{( t )} - \eta ( \mathbf{A} \mathbf{w}^{( t )} + \mathbf{b} + \boldsymbol{\xi}^{( t )} + \alpha \mathbf{h}^{( t )} + \boldsymbol{\varepsilon}^{( t )} )
\end{equation}
where auxiliary variables are defined as follows:
\begin{equation}
\boldsymbol{\xi}^{( t )} = \frac{1}{P} \sum_{p = 1}^{P} \boldsymbol{\xi}_{p}^{( t )},~ \mathbf{h}^{( t )} = \frac{1}{P} \sum_{p = 1}^{P} \mathbf{h}_{p}^{( t )},~ \boldsymbol{\varepsilon}^{( t )} = \frac{1}{P} \sum_{p = 1}^{P} \boldsymbol{\varepsilon}_{p}^{( t )}
\end{equation}

From Lemma \ref{lem:opt_qsgd_quan_err_bnd}, we can derive the variance bound for the pseudo quantization error:
\begin{equation}
\mathbb{E} \| \boldsymbol{\varepsilon}^{( t )} \|_{2}^{2} \le \left[ 1 + \alpha^{2} \gamma \cdot \frac{1 - \lambda^{t}}{1 - \lambda} \right] \cdot \frac{\gamma B}{P}
\label{eqn:opt_qsgd_psd_quan_err_bnd}
\end{equation}

Below, we present the worst-case upper bound on the expected distance between $\mathbf{w}^{( t + 1 )}$, solution obtained in the $( t + 1 )$-th iteration, and $\mathbf{w}^{*}$, the optimal solution.

\begin{theorem}
Let $f \left( \mathbf{w} \right) = \frac{1}{2} \mathbf{w}^{T} \mathbf{A} \mathbf{w} + \mathbf{b}^{T} \mathbf{w}$ be the objective function to be minimized, whose optimal solution is denoted as $\mathbf{w}^{*}$, and $R^{2} = \sup_{\mathbf{w} \in \mathbb{R}^{d}} \| \mathbf{w} - \mathbf{w}^{*} \|_{2}^{2}$. Assume the stochastic noise in the mini-batch's gradients satisfies $\mathbb{E} \| \boldsymbol{\xi}^{( t )} \|_{2}^{2} \le \sigma^{2}$, then the error bound in the ECQ-SGD algorithm satisfies:
\begin{equation}
\begin{split}
\mathbb{E} \| \mathbf{w}^{( t + 1 )} - \mathbf{w}^{*} \|_{2}^{2} &\le R^{2} \| \mathbf{H}^{t + 1} \|_{2}^{2} + \eta^{2} \sigma^{2} \sum_{t' = 0}^{t} \| \mathbf{H}^{t'} \|_{2}^{2} \\
+ \eta^{2} \mathbb{E} \| \boldsymbol{\varepsilon}^{( t )} \|_{2}^{2} &+ \eta^{2} \sum_{t' = 0}^{t - 1} \| \boldsymbol{\Theta}^{( t' )} \|_{2}^{2} \cdot \mathbb{E} \| \boldsymbol{\varepsilon}^{( t' )} \|_{2}^{2}
\label{eqn:opt_qsgd_error_bound}
\end{split}
\end{equation}
where $\mathbf{H} = \mathbf{I} - \eta \mathbf{A}$ and:
\begin{equation}
\boldsymbol{\Theta}^{( t' )} = \mathbf{H}^{t - t'} - \sum_{t'' = t' + 1}^{t} \alpha ( \beta - \alpha )^{t'' - t' - 1} \mathbf{H}^{t - t''}
\end{equation}
\end{theorem}

\begin{proof}
Please refer to the supplementary material.
\end{proof}

It is worth mentioning that our algorithm only differs from QSGD in the last two terms of error bound given above; the other two terms are identical in both algorithms. Now we analyse $\boldsymbol{\Theta}^{( t' )}$'s value to reveal the difference between ECQ-SGD and QSGD. Firstly, we can prove that:
\begin{lemma}
If the learning rate satisfies $\eta a_{1} < 1$, then:
\begin{equation}
\boldsymbol{\Theta}^{( t' )} \preceq \left( 1 - \frac{\alpha}{1 - \eta a_{1}} \frac{1 - \nu^{t - t'}}{1 - \nu} \right) \cdot \mathbf{H}^{t - t'}
\end{equation}
where $\nu = ( \beta - \alpha) / ( 1 - \eta a_{1} )$.
\label{lem:theta_bound}
\end{lemma}

By letting $0 < \alpha < 1$ and $\beta = 1 - \eta a_{1}$, the inequality in Lemma \ref{lem:theta_bound} is simplified to $\boldsymbol{\Theta}^{( t' )} \preceq \nu^{t - t'} \cdot \mathbf{H}^{t - t'}$. Also, we have $0 < \nu < 1$ from its definition, which implies that in ECQ-SGD, the multiplier of each previous quantization error consistently precedes its counterpart in QSGD (where $\nu = 1$ due to $\alpha = 0$).

To simplify further discussions, we introduce the auxiliary variable $\tau^{( t' )}$, defined as:
\begin{equation}
\tau^{( t' )} = \sup \left( \| \boldsymbol{\Theta}^{( t' )} \|_{2}^{2} \cdot \mathbb{E} \| \boldsymbol{\varepsilon}^{( t' )} \|_{2}^{2} \right)
\end{equation}
which stands for the upper bound of quantization error $\boldsymbol{\varepsilon}^{( t' )}$'s contribution to the final error bound (\ref{eqn:opt_qsgd_error_bound}). Similarly, the upper bound of quantization error's contribution in QSGD is given by:
\begin{equation}
\tau_{QSGD}^{( t' )} = \| \mathbf{H}^{t - t'} \|_{2}^{2} \cdot \frac{\gamma B}{P}
\end{equation}
which is actually a special case of ECQ-SGD when $\alpha = 0$.

Recall that in (\ref{eqn:simple_variance_bound_comp}), by ensuring $\lambda < 1$, the variance bound of quantization error in ECQ-SGD is relatively close to that of QSGD. Therefore, with properly chosen hyper-parameters, ECQ-SGD's reduction ratio in $\tau^{( t' )}$ (comparing against QSGD) approaches to zero as the time gap $\left( t - t' \right)$ grows to infinity:
\begin{lemma}
If $\beta = 1 - \eta a_{1}$ and $\alpha > 0$ satisfies that $\lambda = \alpha^{2} \gamma + ( \beta - \alpha )^{2} < 1$, then we have:
\begin{equation}
\lim_{\left( t - t' \right) \rightarrow \infty} \frac{\tau^{( t' )}}{\tau_{QSGD}^{( t' )}} = 0
\end{equation}
\label{lem:err_bound_reduction}
\end{lemma}

The detailed proof for Lemma \ref{lem:theta_bound} and \ref{lem:err_bound_reduction} can be found in the supplementary material. In practice, since the learning rate $\eta \ll \frac{1}{a_{1}}$, we can simply set $\beta = 1$ and $\alpha$ to some small positive number to approximately satisfy the requirements in Lemma \ref{lem:err_bound_reduction}.

Lemma \ref{lem:err_bound_reduction} implies that as the iteration goes on, ECQ-SGD can better suppress all the previous quantization errors' contribution to the error bound (\ref{eqn:opt_qsgd_error_bound}) than QSGD. An intrinsic understanding is that with the error feedback scheme, quantization errors from different iterations are cancelled out when evaluating the final error bound, leading to a tighter worst-case upper bound than that of QSGD.

\begin{figure*}[!ht]
\begin{center}
\centerline{
	\includegraphics[trim={0 0 0.5in 0.5in}, clip, width=.56\columnwidth]{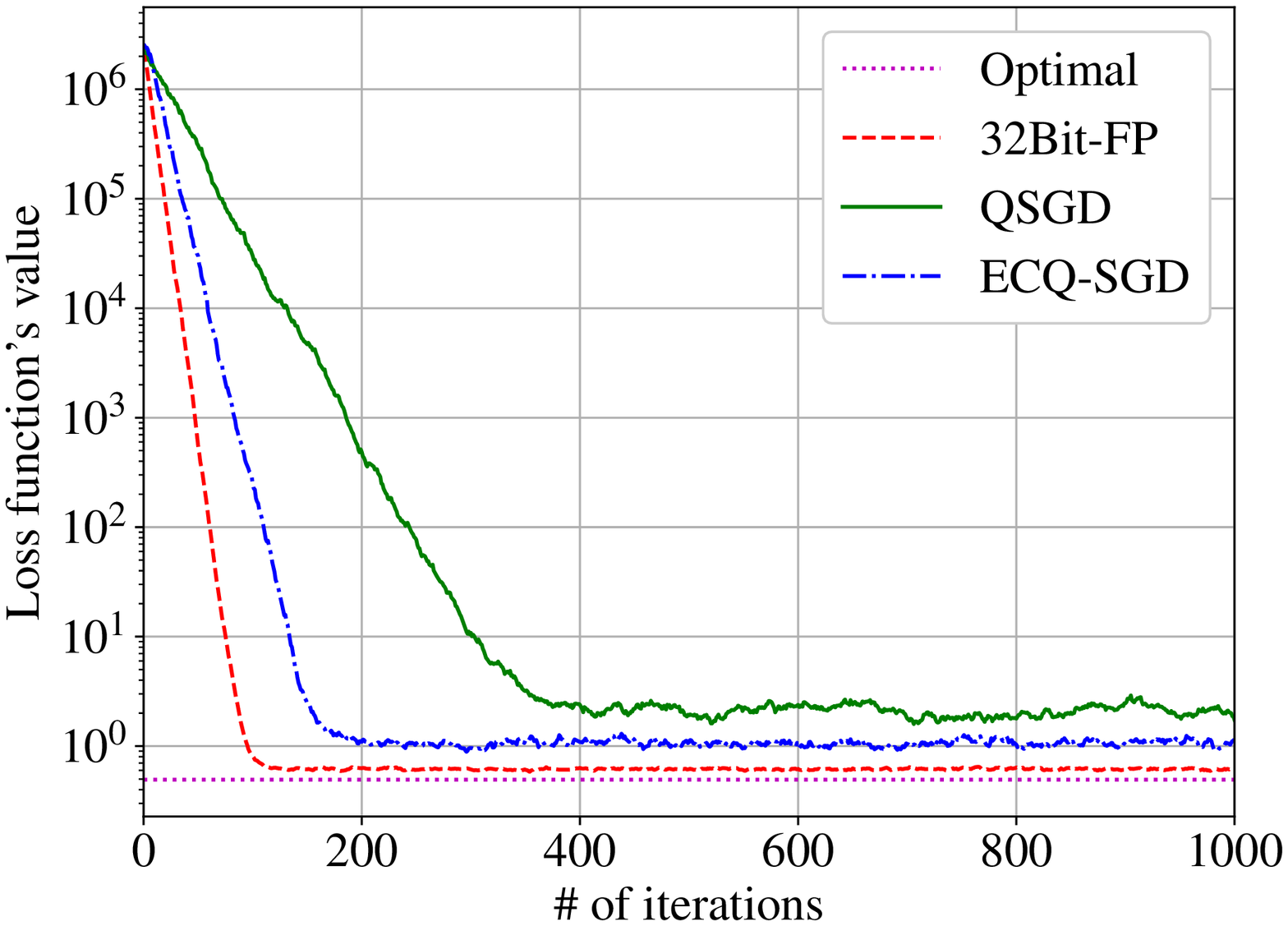} \hspace{0.2in}
	\includegraphics[trim={0 0 0.5in 0.5in}, clip, width=.56\columnwidth]{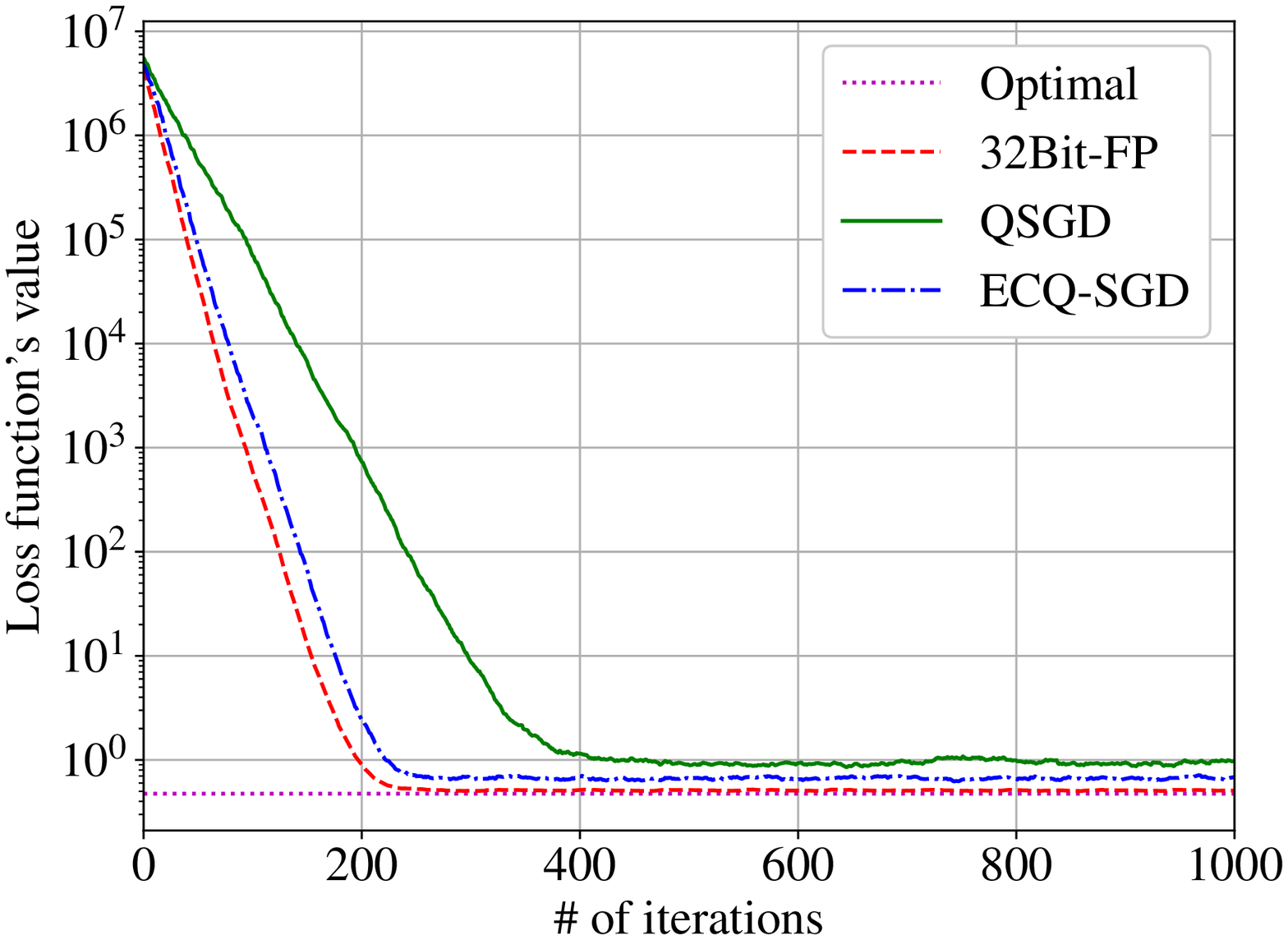} \hspace{0.2in}
	\includegraphics[trim={0 0 0.5in 0.5in}, clip, width=.56\columnwidth]{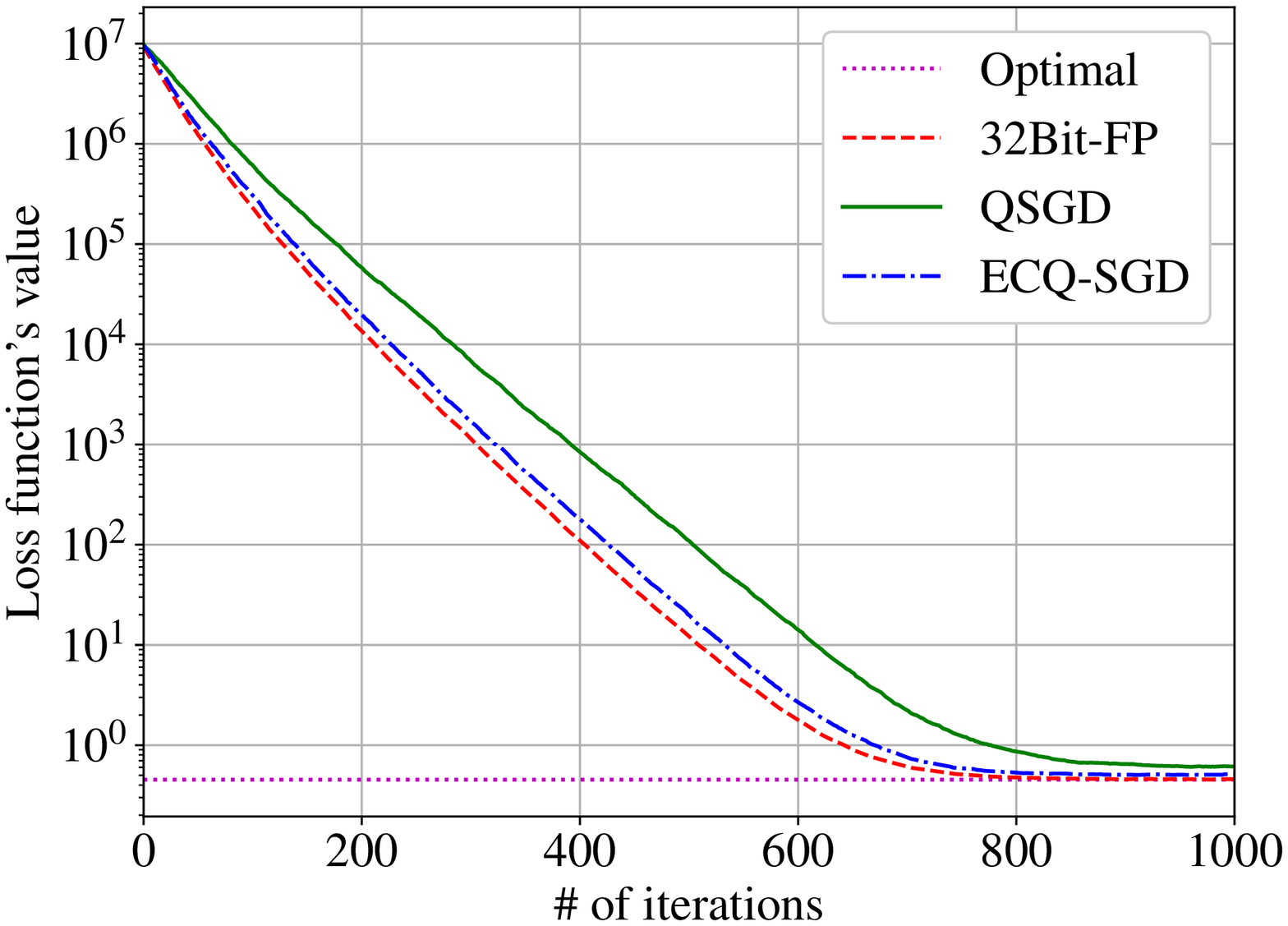}
}
\centerline{
	\includegraphics[trim={0 0 0.5in 0.5in}, clip, width=.56\columnwidth]{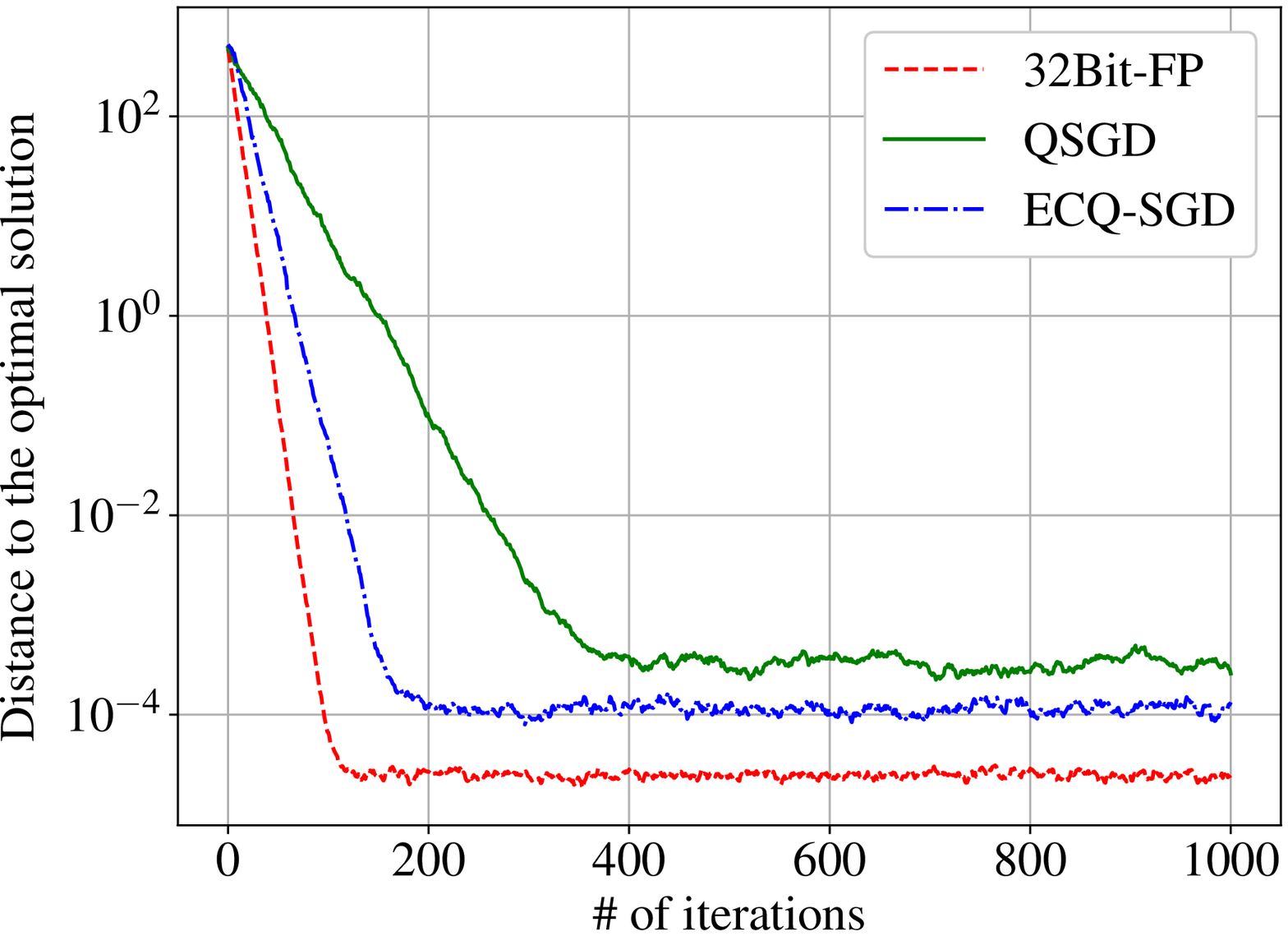} \hspace{0.2in}
	\includegraphics[trim={0 0 0.5in 0.5in}, clip, width=.56\columnwidth]{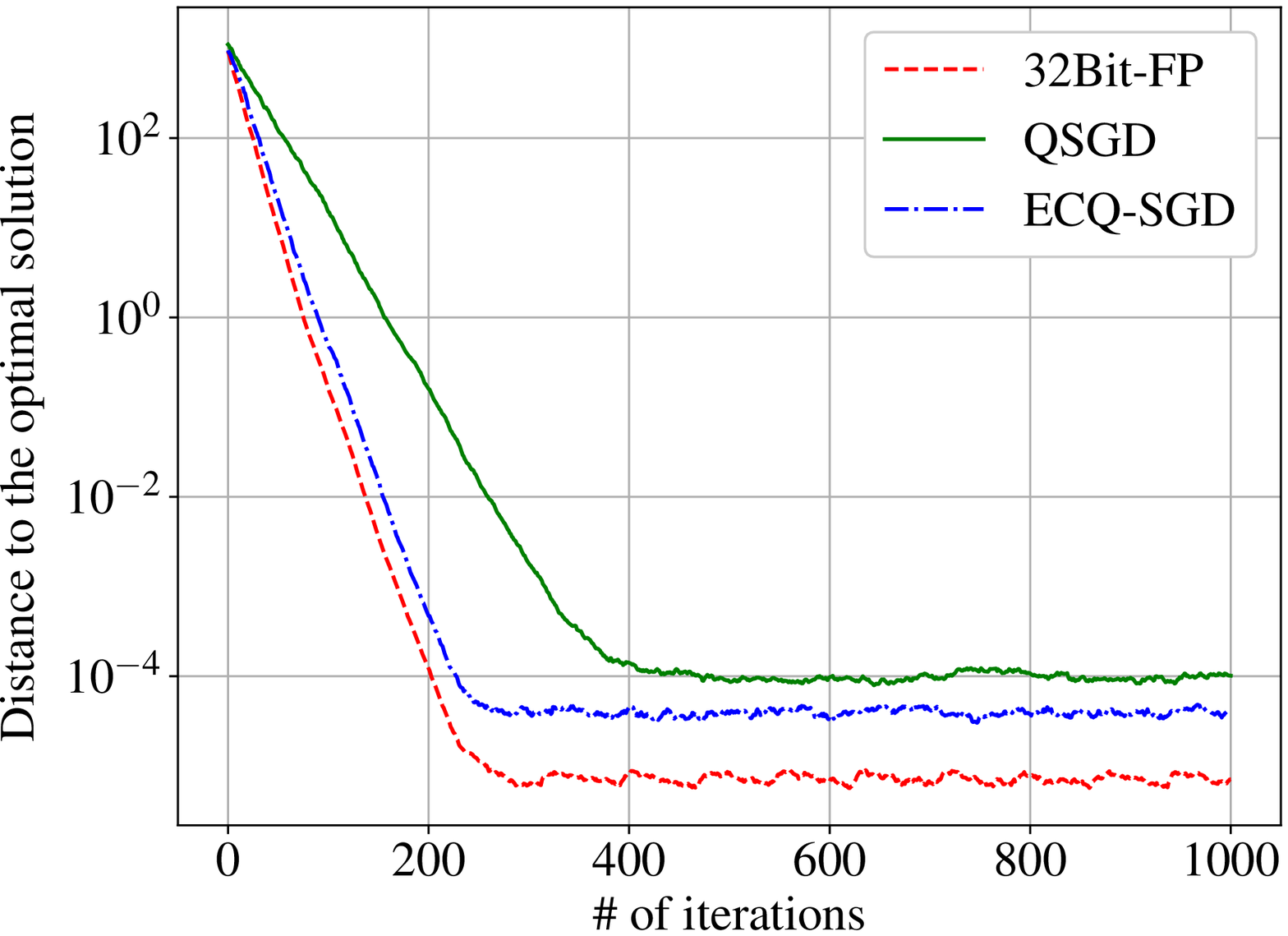} \hspace{0.2in}
	\includegraphics[trim={0 0 0.5in 0.5in}, clip, width=.56\columnwidth]{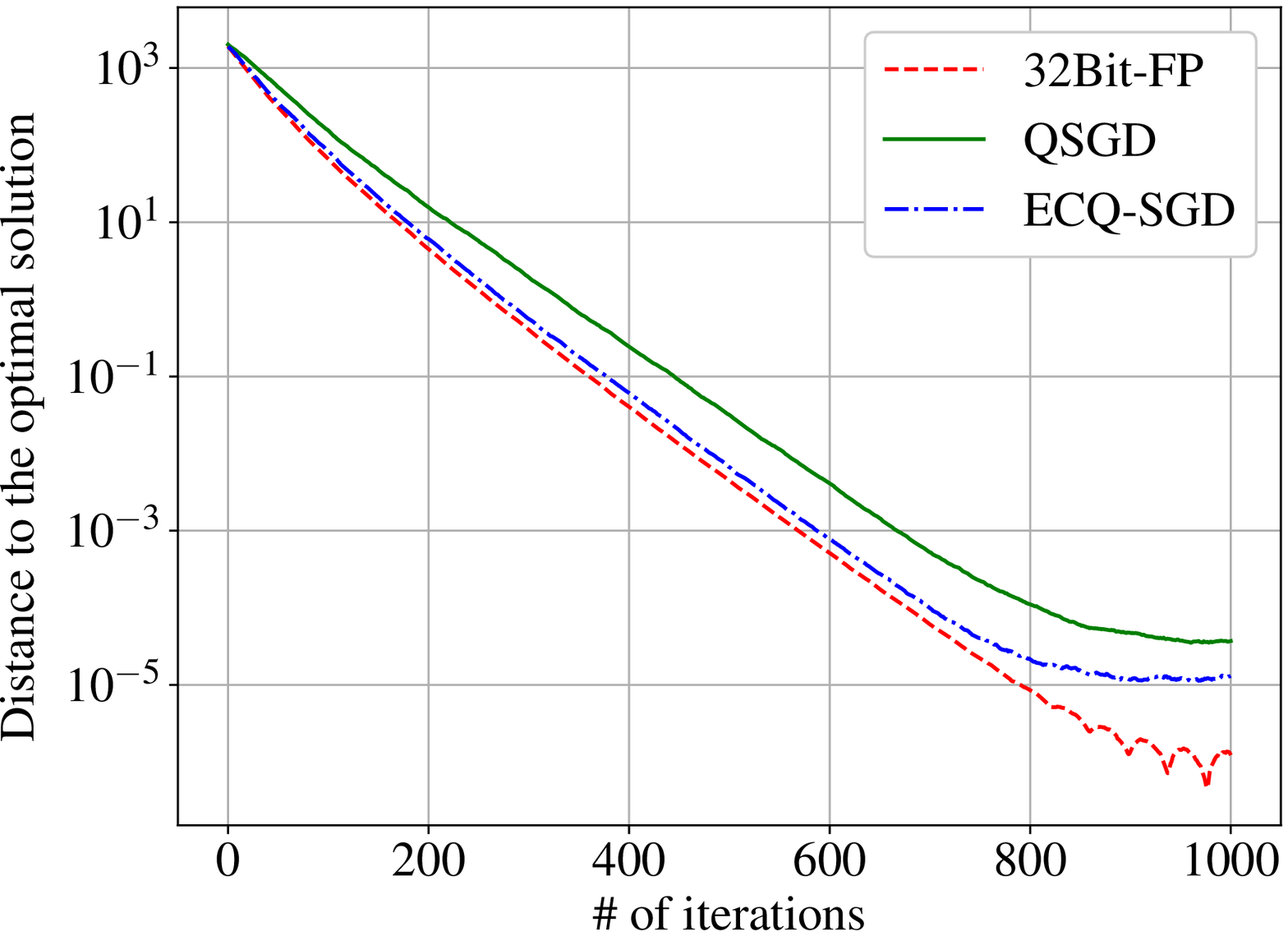}
}
\caption{Comparison on the loss function's value and distance to the optimal solution (left: \textit{Syn-256}; middle: \textit{Syn-512}; right: \textit{Syn-1024}).}
\label{fig:loss_n_err_bound_syn}
\end{center}
\end{figure*}

\section{Experiments}

In this section, we demonstrate the effectiveness of our proposed ECQ-SGD algorithm with extensive experiments. We start with linear models for convex optimization, and then extend to non-convex optimization with deep convolutional neural networks. We further analyse the scalability for large-scale scenarios, and finish the discussion with a detailed study on choices of hyper-parameters.

\subsection{Linear Models}

In the previous convergence analysis, we claim that for convex quadratic optimization, ECQ-SGD can reduce the quantization error's contribution to the error bound, which leads to better solution after convergence. Now we verify this by comparing the convergence behaviour of ECQ-SGD and QSGD for training linear regression models.

Here we start with three synthetic datasets: Syn-256, Syn-512, and Syn-1024. Each dataset consists of 10k training samples, and the suffix denotes the feature dimension $d$. The training sample is generated as $y_{i} = \mathbf{w}^{*T} \mathbf{x}_{i} + \epsilon_{i}$, where $\mathbf{w}^{*} \in \mathbb{R}^{d}$ is the underlying model parameters we wish to obtain, and $\{ \epsilon_{i} \}$ are \iid random noises. The learning rate is set to 0.02, and both QSGD and ECQ-SGD use $l_{2}$-norm as the scaling factor and 4 non-zero quantization levels, \ie $s = 4$. For ECQ-SGD, we let $\alpha = 0.2$ and $\beta = 0.9$.

In Figure \ref{fig:loss_n_err_bound_syn}, we compare the loss function's value (top) and distance to the optimal solution (bottom) in each iteration. For all three datasets, the convergence of loss function's value of ECQ-SGD is more close to the 32-bit full-precision SGD, and significantly faster than QSGD. On the other hand, the gap between QSGD (or ECQ-SGD) and 32Bit-FP in the distance to the optimal solution measures the quantization error's contribution to the error bound as defined in (\ref{eqn:opt_qsgd_error_bound}). The distance gap of ECQ-SGD is clearly smaller than QSGD, indicating that the quantization error's contribution to the error bound is well suppressed.

\begin{figure}[ht]
\begin{center}
\centerline{\includegraphics[width=.9\columnwidth]{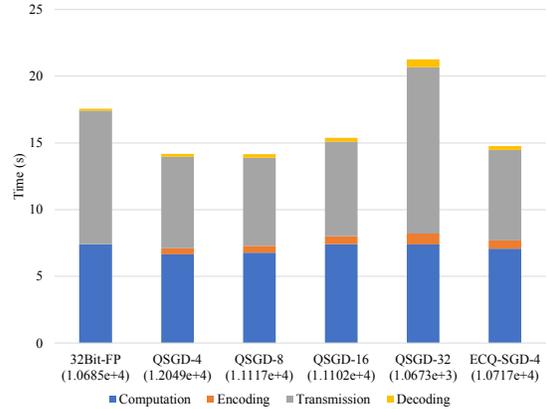}}
\caption{Comparison on the decomposed time consumption and test loss (in brackets) on the Syn-20K dataset. The suffix in QSGD and ECQ-SGD represents $s$, the number of non-zero quantization levels. The time consumption is for 1k iterations in total.}
\label{fig:time_consumption_syn_20k}
\end{center}
\end{figure}

Now we compare the run-time speed between QSGD and ECQ-SGD on a larger dataset, Syn-20k, which consists of 50k training samples and the feature dimension is 20k. In Figure \ref{fig:time_consumption_syn_20k}, we report the decomposed time consumption and test loss (in brackets) of various methods after 1k iterations. We discover that ECQ-SGD achieves similar test loss with 32Bit-FP in a shorter time than both 32Bit-FP and QSGD. Although ECQ-SGD requires extra encoding and decoding time, the overall training speed is still improved due to the reduction in the gradient communication time.

\begin{figure}[!ht]
\centering
\includegraphics[trim={0 0 0.5in 0.5in}, clip, width=.48\columnwidth]{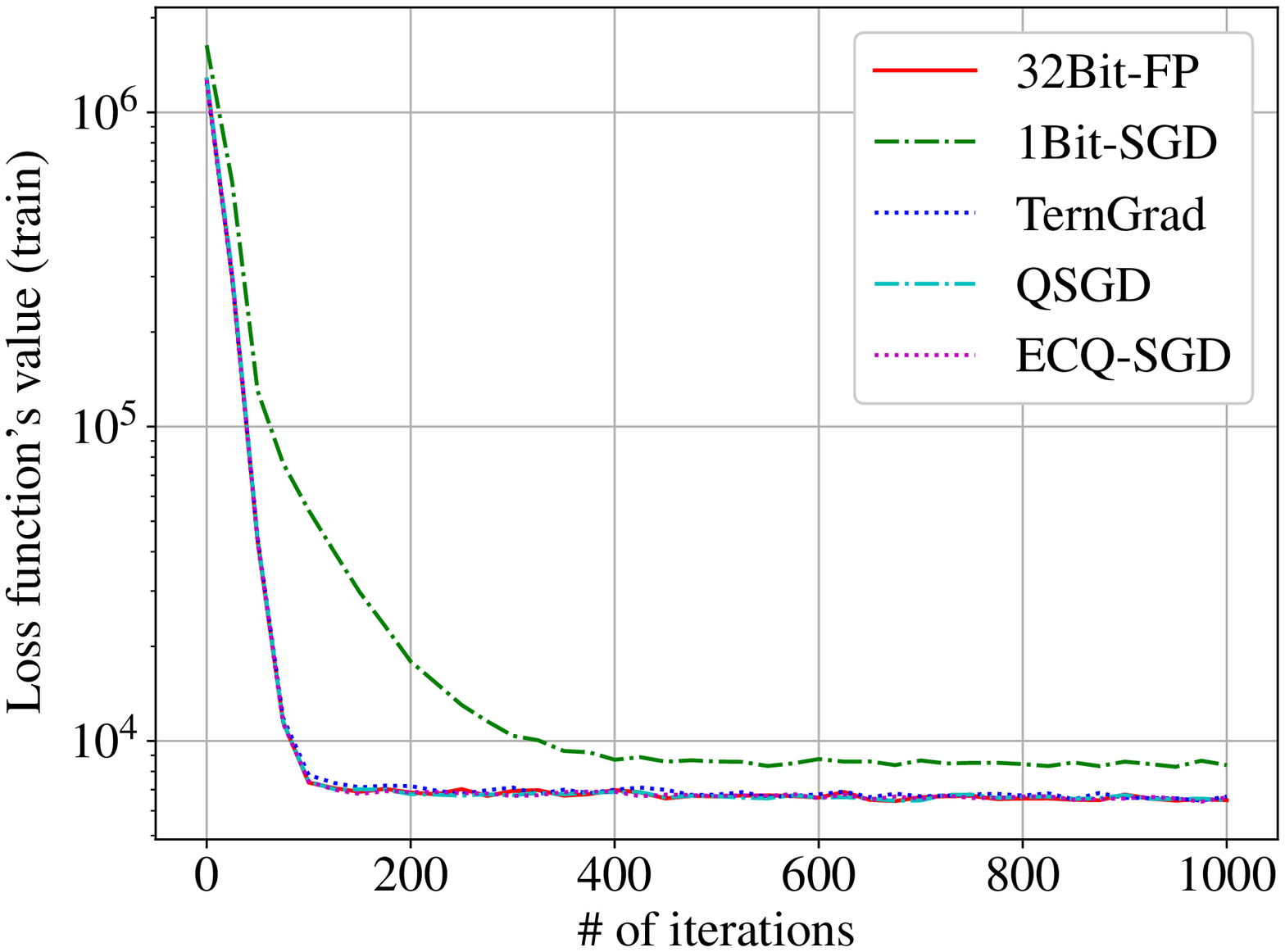}\xspace
\includegraphics[trim={0 0 0.5in 0.5in}, clip, width=.48\columnwidth]{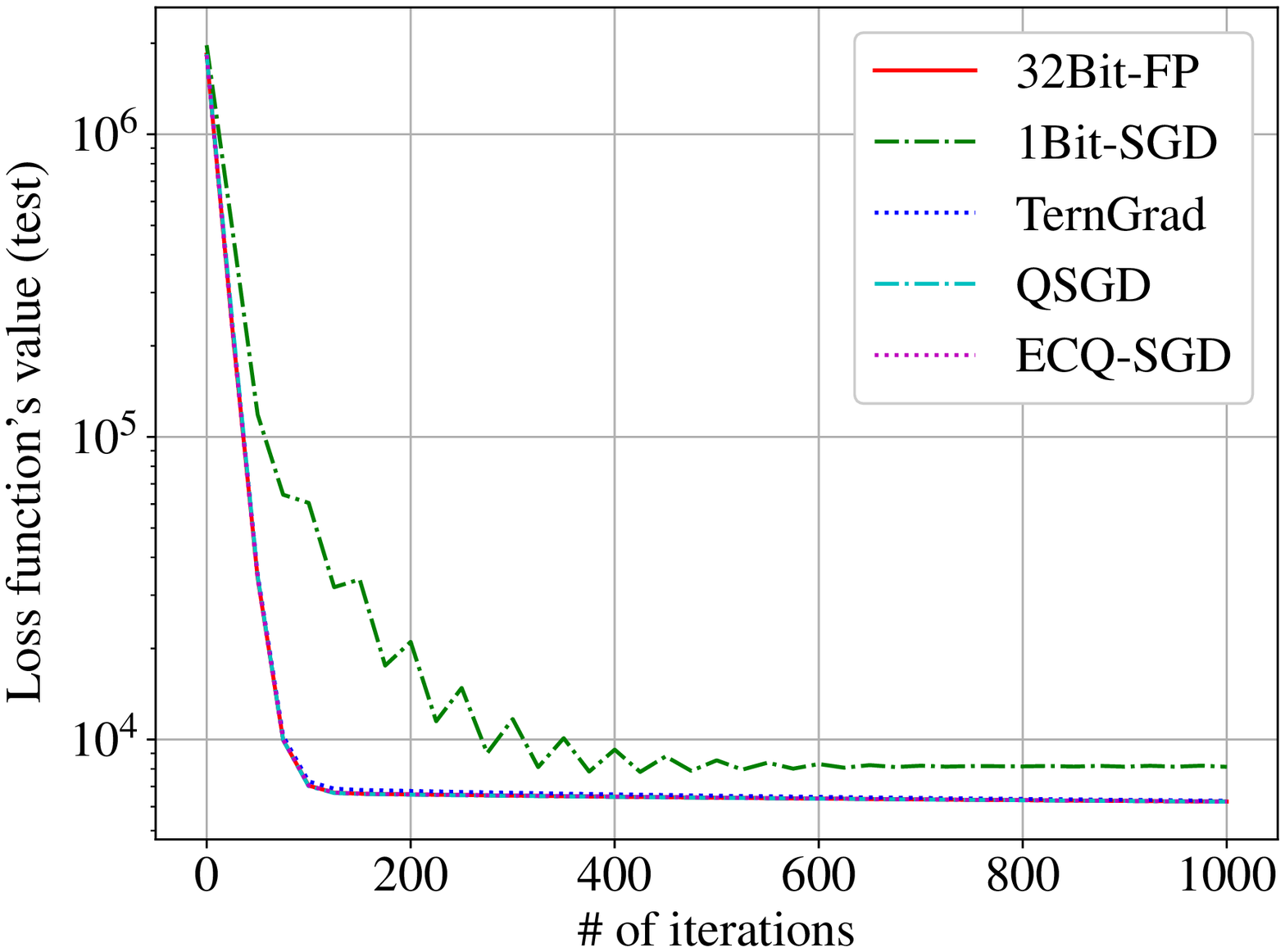}
\includegraphics[trim={0 0 0.5in 0.5in}, clip, width=.48\columnwidth]{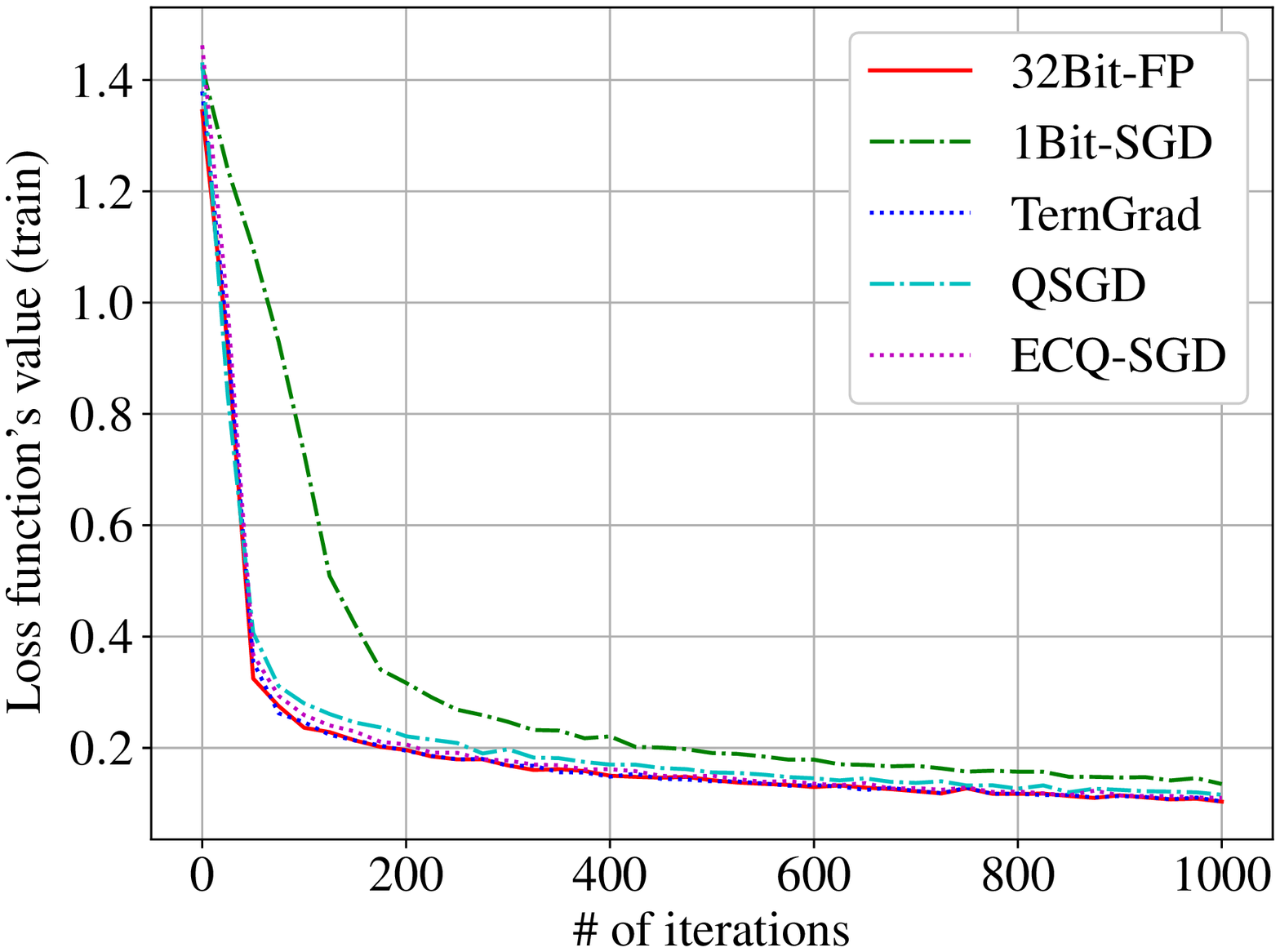}\xspace
\includegraphics[trim={0 0 0.5in 0.5in}, clip, width=.48\columnwidth]{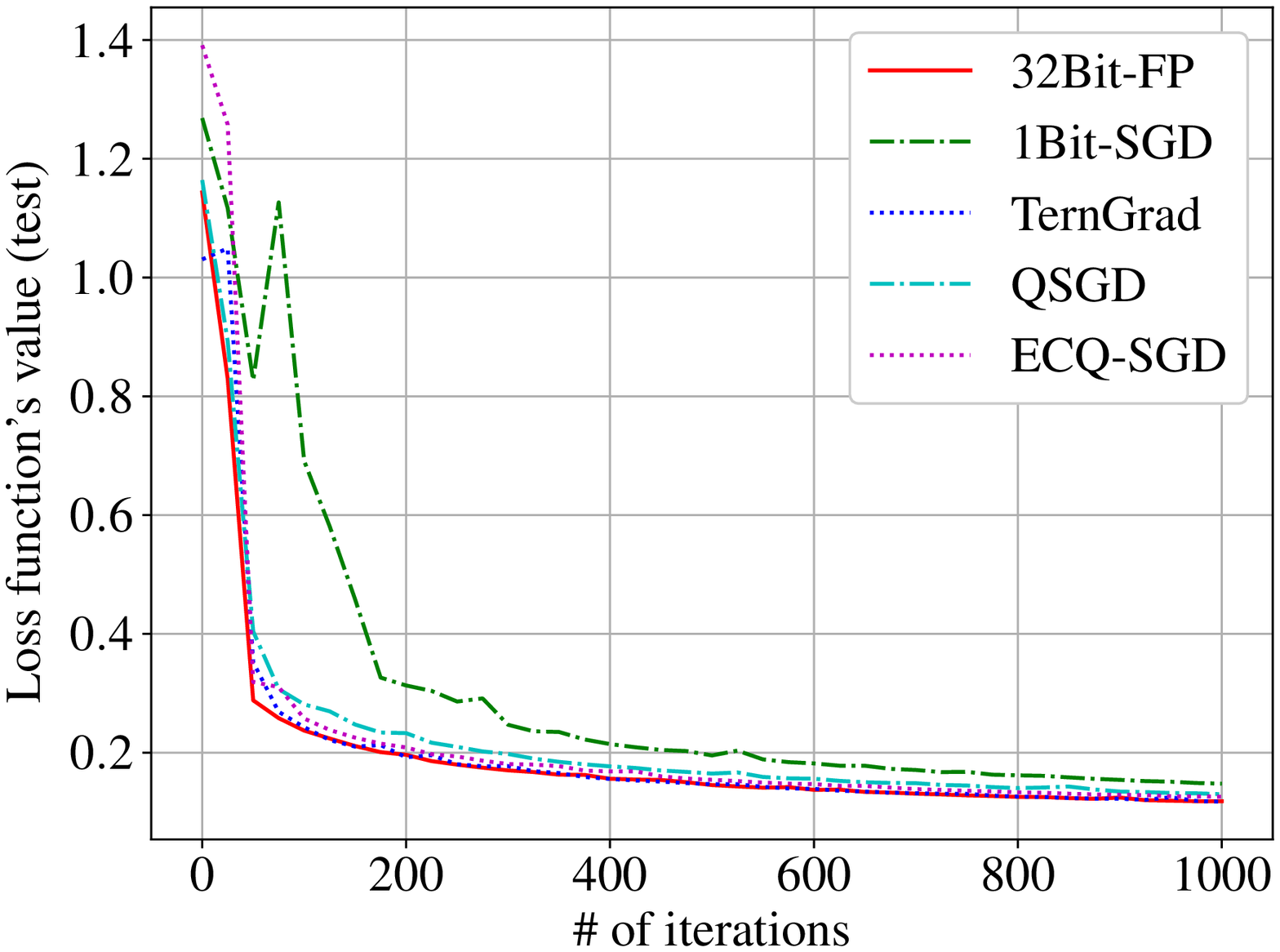}
\caption{Comparison on the loss function's value for training linear models with various methods (top: \textit{YearPredictionMSD}; bottom: \textit{gisette}; left: training loss; right: test loss).}
\label{fig:loss_vs_iter_year_gisette}
\end{figure}

Furthermore, we extend the evaluation to two publicly available datasets, \textit{YearPredictionMSD} for regression and \textit{gisette} for classification \cite{chang2011libsvm}. Linear regression and logistic regression models are trained with different gradient quantization methods on these two datasets respectively. The comparison on the loss function's value and communication cost, averaged from five random runs, is as depicted in Figure \ref{fig:loss_vs_iter_year_gisette} and Table \ref{tab:loss_comm_year_gisette}. Here we use squared $l_{2}$-loss for regression and log-loss for classification. The communication cost is measured by the total number of bits to encode gradients after 1k iterations. We use entropy encoding to fully exploit the sparsity after gradient quantization for all methods.

\begin{table}[!ht]
\caption{Comparison on the loss function's value and overall communication cost for training linear models with various methods.}
\label{tab:loss_comm_year_gisette}
\centering
\begin{small}
\begin{sc}
\begin{tabular}{lccc}
\toprule
\scriptsize{YearPredictionMSD} & Loss & \# of Bits & Ratio \\
\midrule
32Bit-FP & $\mathbf{4.99e{3}}$ & $2.88e{6}$ & - \\
1Bit-SGD & $6.57e{3}$ & $1.51e{5}$ & $19.07\times$ \\
TernGrad & $5.04e{3}$ & $1.41e{5}$ & $20.43\times$ \\
QSGD     & $5.11e{3}$ & $\mathbf{1.17e{5}}$ & $\mathbf{24.63\times}$ \\
ECQ-SGD  & $5.00e{3}$ & $1.22e{5}$ & $23.62\times$ \\
\midrule
Gisette & Loss & \# of Bits & Ratio \\
\midrule
32Bit-FP & $\mathbf{1.16e{-1}}$ & $1.60e{8}$ & - \\
1Bit-SGD & $1.47e{-1}$ & $2.59e{6}$ & $61.75\times$ \\
TernGrad & $\mathbf{1.16e{-1}}$ & $4.62e{6}$ & $34.67\times$ \\
QSGD     & $1.48e{-1}$ & $5.88e{5}$ & $272.18\times$ \\
ECQ-SGD  & $\mathbf{1.16e{-1}}$ & $\mathbf{5.68e{5}}$ & $\mathbf{281.88\times}$ \\
\bottomrule
\end{tabular}
\end{sc}
\end{small}
\end{table}

From Figure \ref{fig:loss_vs_iter_year_gisette}, we observe that all methods except 1Bit-SGD converge at similar speed, and the final performance is also almost identical to each other. This may due to the relatively small number of feature dimensions (90) of the \textit{YearPredictionMSD} dataset, so that different gradient quantization methods do not have significant performance gap.

For the \textit{gisette} dataset, whose feature dimension is 5000, both TernGrad and ECQ-SGD can still match the performance of the 32-bit full-precision SGD, while 1Bit-SGD and QSGD suffer more severe performance degradation. On the other hand, ECQ-SGD achieves much higher compression ratio ($281.88\times$) than TernGrad ($34.67\times$).

\subsection{Convolutional Neural Networks}

Now we evaluate the ECQ-SGD algorithm for training convolutional neural networks, which is a highly non-convex optimization problem.

\begin{figure*}[!ht]
\begin{center}
\centerline{
	\includegraphics[trim={0 0 0.5in 0.5in}, clip, width=.56\columnwidth]{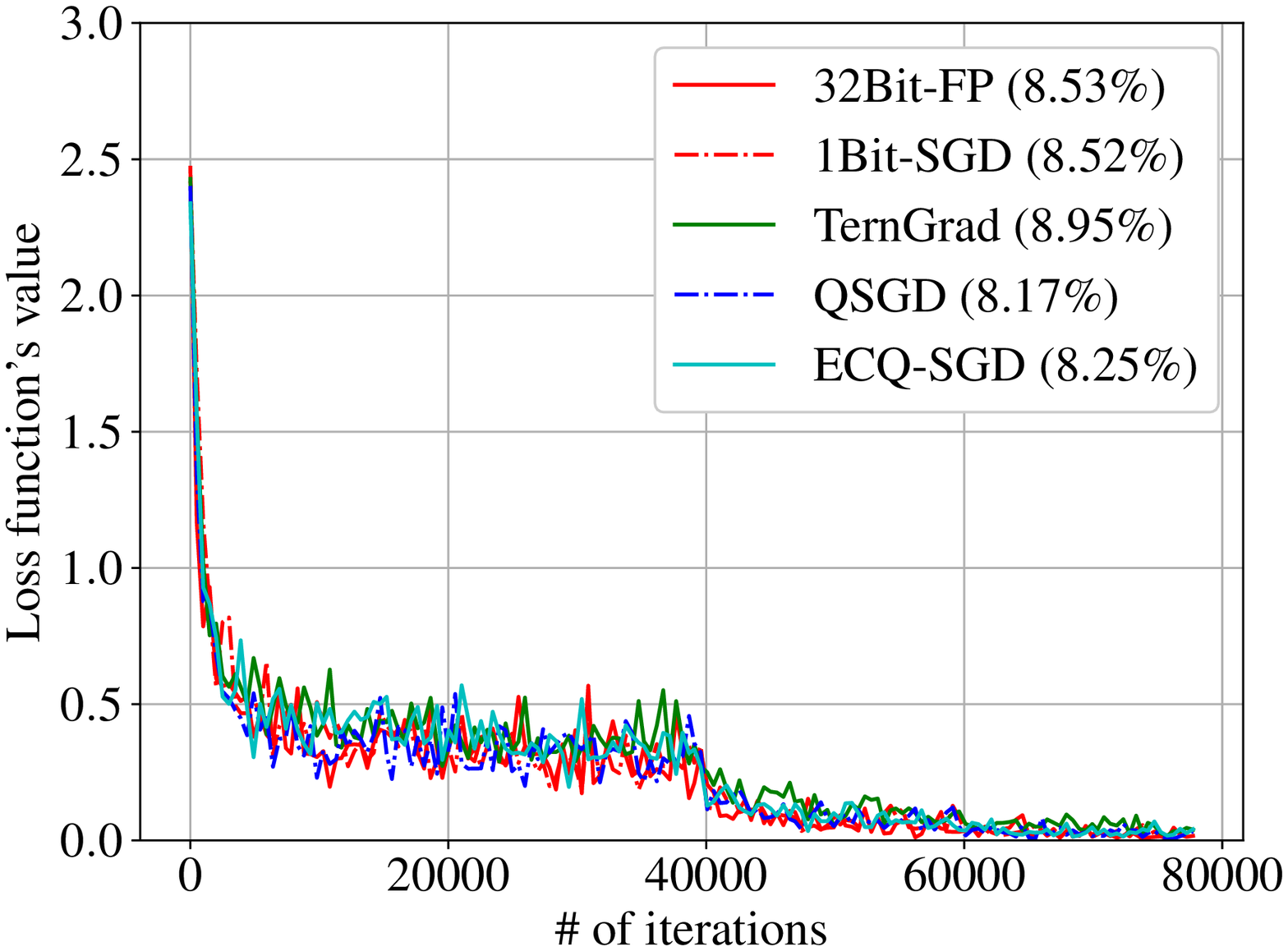} \hspace{0.2in}
	\includegraphics[trim={0 0 0.5in 0.5in}, clip, width=.56\columnwidth]{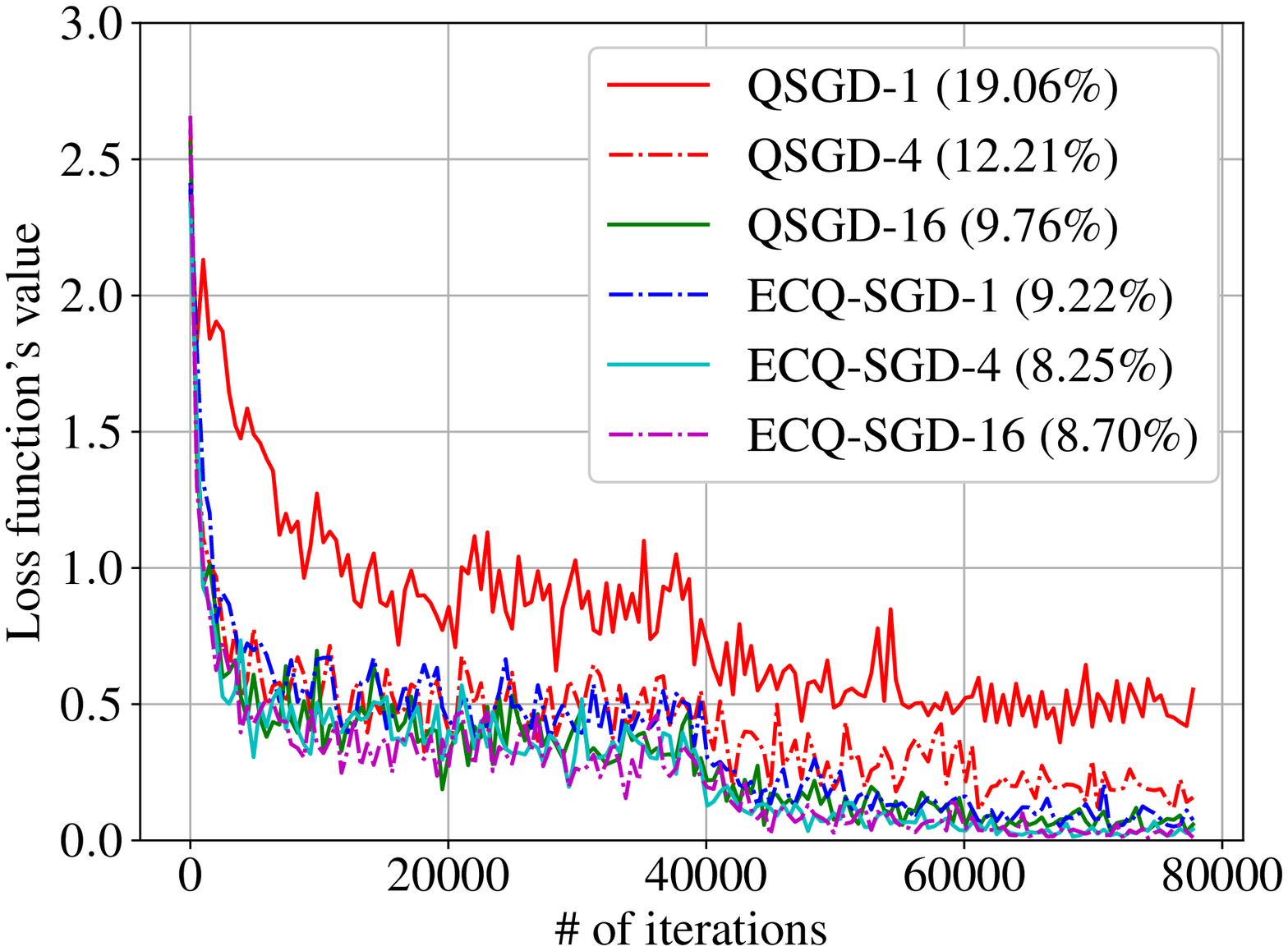} \hspace{0.2in}
	\includegraphics[trim={0 0 0.5in 0.5in}, clip, width=.56\columnwidth]{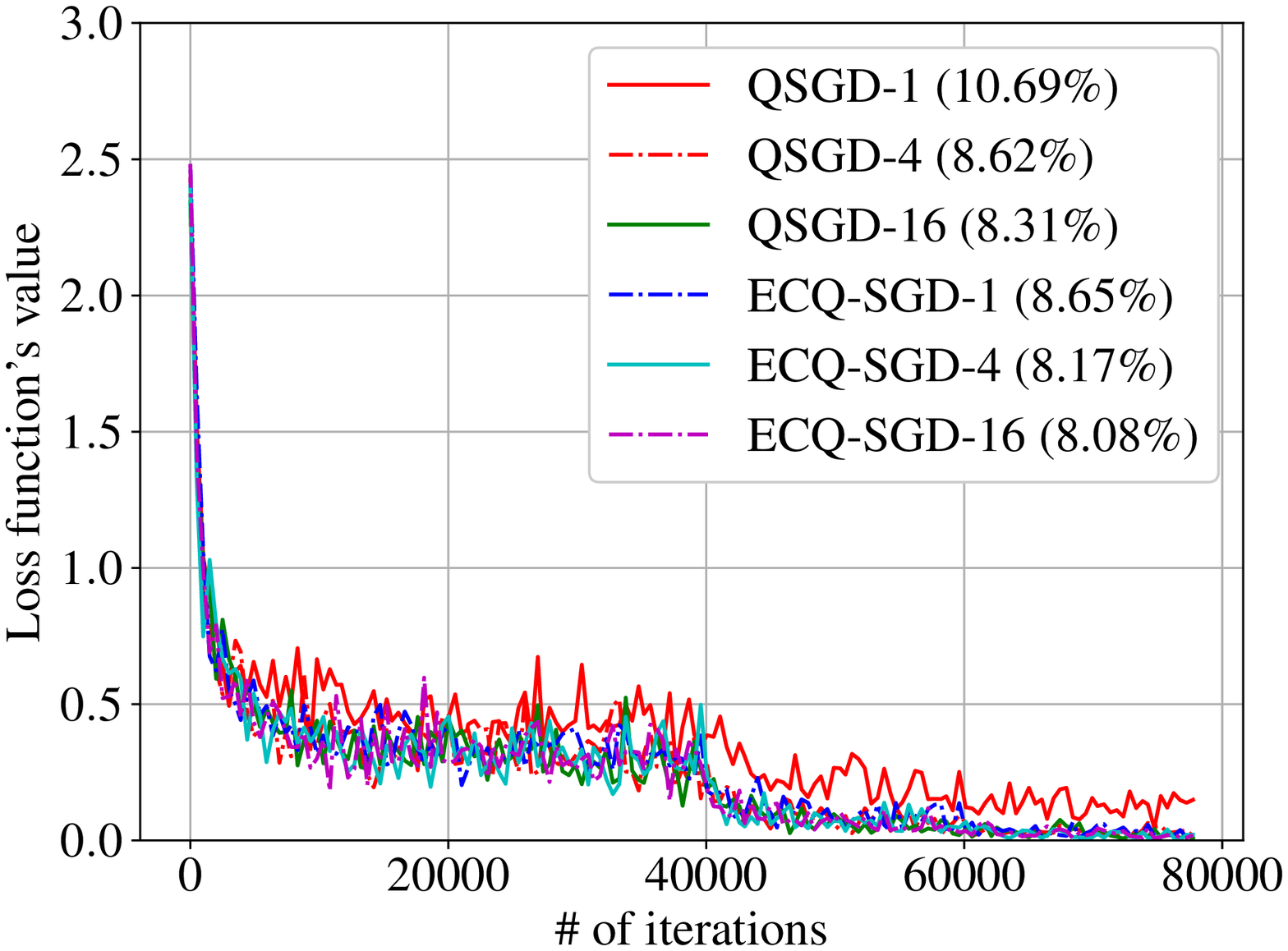}
}
\centerline{
	\includegraphics[trim={0 0.3in 0 0.2in}, clip, width=.56\columnwidth]{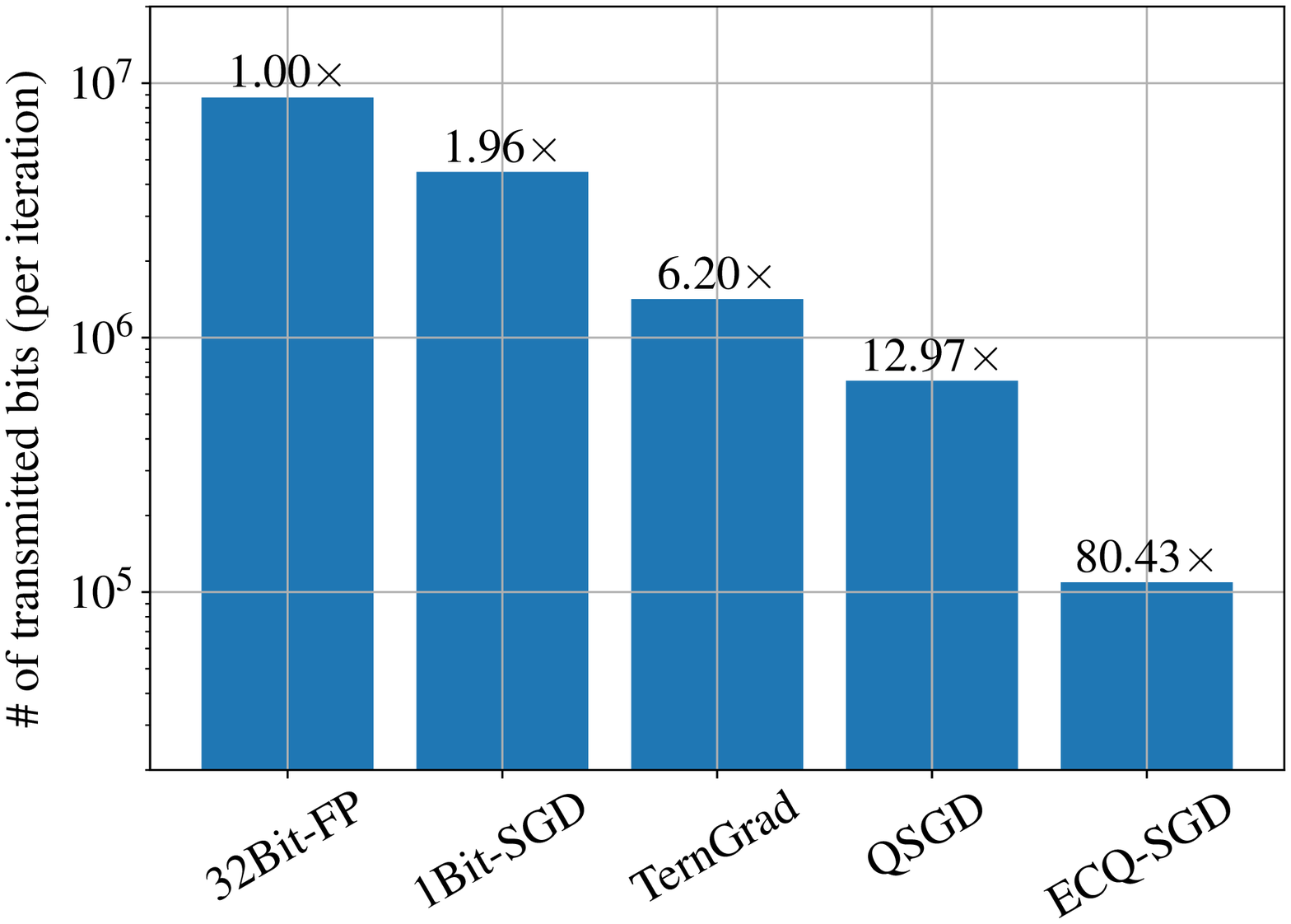} \hspace{0.2in}
	\includegraphics[trim={0 0.3in 0 0.2in}, clip, width=.56\columnwidth]{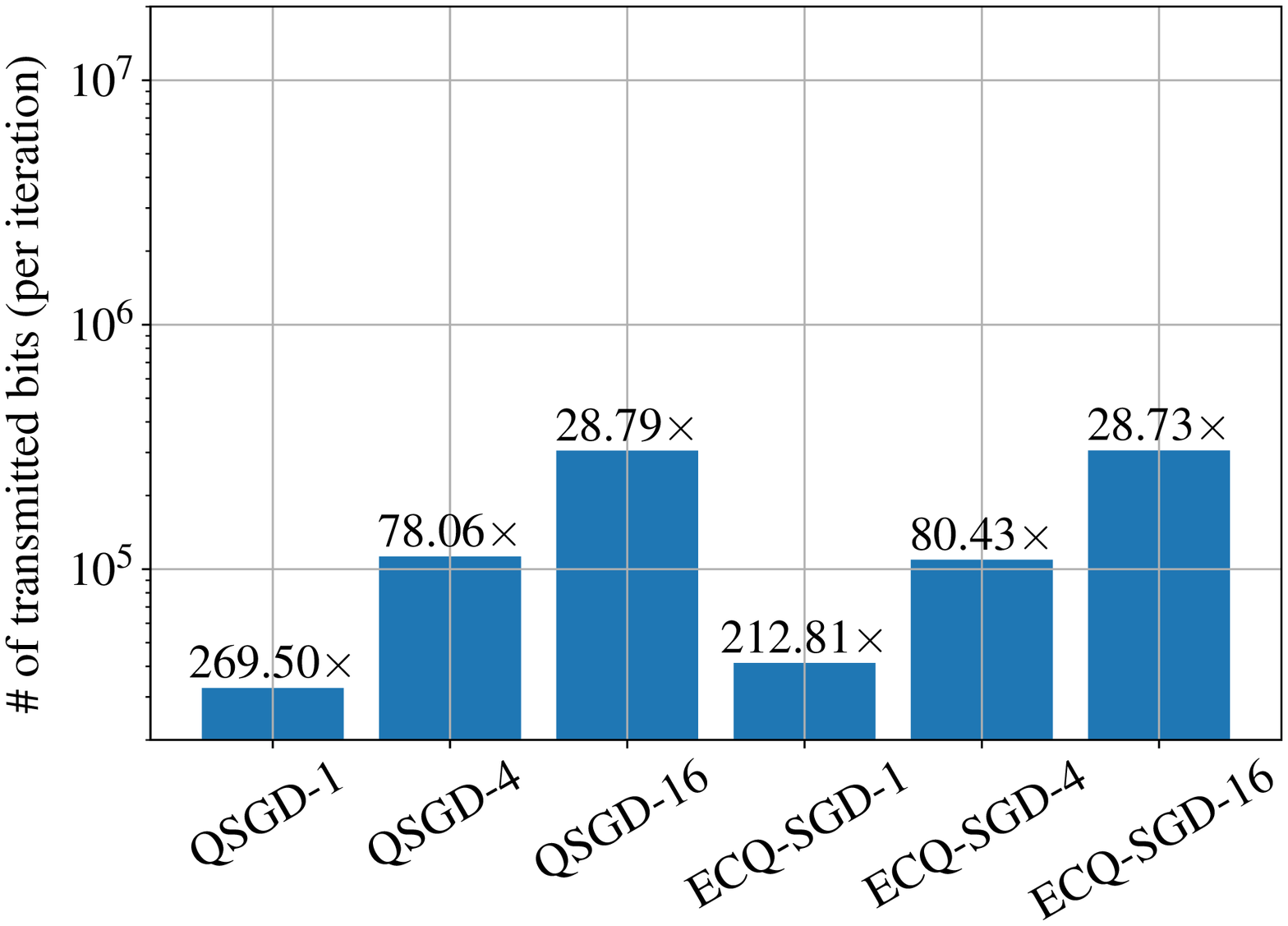} \hspace{0.2in}
	\includegraphics[trim={0 0.3in 0 0.2in}, clip, width=.56\columnwidth]{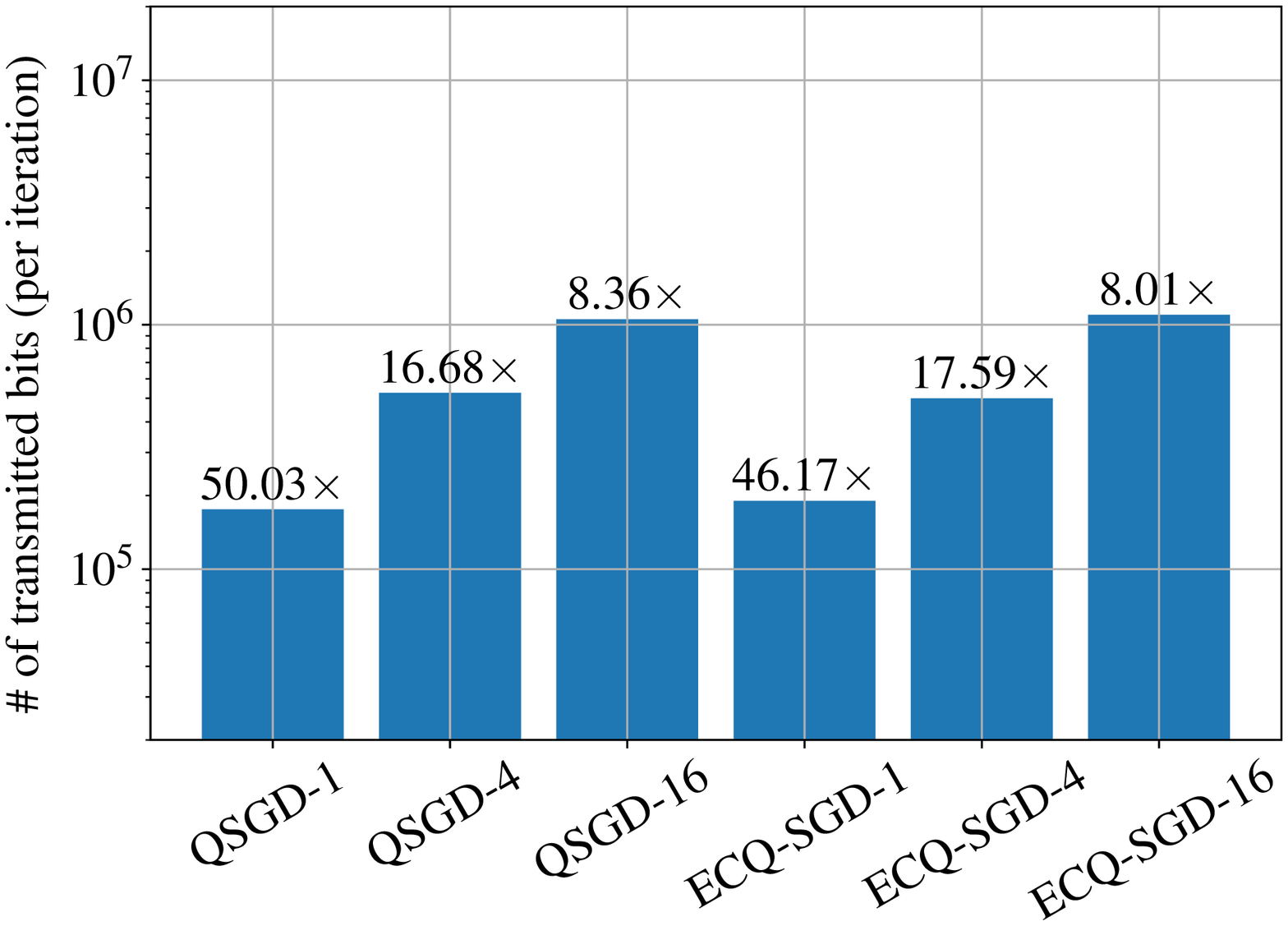}
}
\caption{Comparison on the loss function's value and communication cost (measured by the number of transmitted bits) for training a LeNet model on the CIFAR-10 dataset. The classification accuracy is noted in the bracket. For the first two columns, both QSGD and ECQ-SGD use $l_{\infty}$-norm as the scaling factor, and for the last column, $l_{2}$-norm is used.}
\label{fig:loss_n_comm_cost_cifar10}
\end{center}
\end{figure*}

The experiments are carried out on the CIFAR-10 dataset \cite{krizhevsky2009learning}, which consists of 60k images from 10 categories. We follow the common protocol, using 50k images for training and the remaining 10k images for evaluation. We train the ResNet-20 model \cite{he2016deep} with different gradient quantization methods, and the results are as reported in Figure \ref{fig:loss_n_comm_cost_cifar10}. For all methods, the batch size is set to 128, and the learning rate starts from 0.1, divided by 10 at 40k and 60k iterations. The training process is terminated at the end of the 200-th epoch ($\sim$78k iterations).

In the first column of Figure \ref{fig:loss_n_comm_cost_cifar10}, we compare the loss function's value and overall communication overhead of various methods. The hyper-parameters of each method\footnotemark{} are selected to achieve negligible accuracy loss, comparing with the 32-bit full-precision baseline. We discover that all methods converge at a similar speed, but ECQ-SGD offers over $80\times$ reduction in the communication cost and significantly outperforms other gradient quantization methods.

\footnotetext{For 1Bit-SGD, the bucket size is set to 3. For TernGrad, the bucket size is set to 16. For QSGD, the bucket size is set to 512, using $l_{\infty}$-norm as the scaling factor and $s = 4$. For ECQ-SGD, the bucket size is set to 4096, using $l_{2}$-norm as the scaling factor, and $s = 4$, $\alpha = 0.01$, and $\beta = 1.0$.}

In the second and third column of Figure \ref{fig:loss_n_comm_cost_cifar10}, we present more detailed comparison against QSGD, since it is the most relevant one with our method. Different scaling factors are used: $l_{2}$-norm for the second column, and $l_{\infty}$-norm for the third column. Both methods use a bucket size of 4096 for fair comparison. We observe that ECQ-SGD is consistently superior to QSGD in both convergence speed and classification accuracy, while the reduction in the communication cost of these two methods are similar under the same hyper-parameter settings.

\subsection{Performance Model}

We adopt the performance model proposed in \cite{yan2015performance} to evaluate the scalability of our ECQ-SGD algorithm. Lightweight profiling on the computation and communication time is carried out to estimate the learning efficiency for larger clusters. Major hardware specifications are as follows: Intel Xeon E5-2680 CPU, Nvidia Tesla P40 GPU (8 units per node), and Mellanox ConnectX-3 Pro network card (40Gb/s connectivity).

\begin{figure}[ht]
\begin{center}
\centerline{\includegraphics[width=.9\columnwidth]{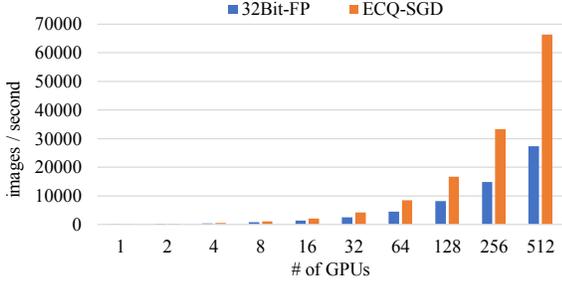}}
\caption{Comparison on the throughput for training a ResNet-50 model on the ILSVRC-12 dataset with different number of GPUs.}
\label{fig:performance_model}
\end{center}
\end{figure}

In Figure \ref{fig:performance_model}, we report the throughput for training a ResNet-50 model on the ILSVRC-12 dataset \cite{russakovsky2015imagenet}. For training with 512 GPUs, ECQ-SGD achieves $143.5\%$ speed-up over the vanilla SGD (66.42k vs. 27.28k images per second). In our experiments, the connection bandwidth is relatively large and thus for clusters with a smaller bandwidth, it is expected that our method can achieve even higher speed-up.

\subsection{Parameter Study}

In Lemma \ref{lem:err_bound_reduction}, we present a guideline for choosing hyper-parameters $\alpha$ and $\beta$ in ECQ-SGD. A simple practice is to set $\beta = 1$ and $\alpha$ to some small positive number satisfying $\alpha^{2} \gamma + ( \beta - \alpha )^{2} < 1$. Now we verify whether this is indeed optimal for training models to higher accuracy.

For quantitative evaluation, we train ResNet-20 models on the CIFAR-10 dataset with various hyper-parameters combinations, and the results are as reported in Table \ref{tab:err_rate_hyper_param}. When fixing $\alpha$ to 0.01, we discover that the lowest error rate is achieved when $\beta$ is set to 1. On the other hand, when $\beta = 1$ is fixed, similar error rate is achieved by setting $\alpha$ to any value between 0.01 and 0.1. If $\alpha$ is too small, then the error feedback effect is greatly weaken, leading to similar performance with QSGD. If $\alpha$ is too large, then the constraint $\lambda = \alpha^{2} \gamma + ( \beta - \alpha )^{2} < 1$ might be violated, and the training process becomes less stable (\eg when $\alpha = 0.15$, the optimization does not converge at all). The above observations are consistent with our previous analysis in Lemma \ref{lem:err_bound_reduction}.

\begin{table}[!ht]
\caption{Comparison on the classification error rate of ResNet-20 on CIFAR-10, under various hyper-parameter combinations.}
\label{tab:err_rate_hyper_param}
\centering
\vskip 0.15in
\begin{small}
\begin{sc}
\begin{tabular}{ccc|ccc}
\toprule
$\alpha$ & $\beta$ & Err. Rate & $\alpha$ & $\beta$ & Err. Rate \\
\midrule
0.01 & 0.90 & 12.05\% & 0.001 & 1.00 & 9.18\% \\
0.01 & 0.95 & 11.47\% & 0.003 & 1.00 & 9.06\% \\
0.01 & 0.98 & 10.69\% & 0.010 & 1.00 & 8.25\% \\
0.01 & 0.99 &  9.64\% & 0.030 & 1.00 & 8.18\% \\
0.01 & 1.00 &  8.25\% & 0.100 & 1.00 & 8.22\% \\
-    & -    & -       & 0.150 & 1.00 & n/a    \\
\bottomrule
\end{tabular}
\end{sc}
\end{small}
\vskip -0.1in
\end{table}

\section{Conclusion}

In this paper, we present the error compensated quantized SGD algorithm to improve the learning efficiency for large-scale distributed optimization. By introducing the error feedback scheme, the ECQ-SGD algorithm can effectively suppress the quantization error's contribution to the error bound. We analyse its convergence behaviour from the theoretical perspective, add demonstrate its advantage over the state-of-the-art QSGD algorithm. Experiments on convex linear models and non-convex convolutional neural networks demonstrate the efficacy of the proposed algorithm.

\bibliography{icml2018_camera_ready_wu}

\begin{thebibliography}{21}
\providecommand{\natexlab}[1]{#1}
\providecommand{\url}[1]{\texttt{#1}}
\expandafter\ifx\csname urlstyle\endcsname\relax
  \providecommand{\doi}[1]{doi: #1}\else
  \providecommand{\doi}{doi: \begingroup \urlstyle{rm}\Url}\fi

\bibitem[Aji \& Heafield(2017)Aji and Heafield]{aji2017sparse}
Aji, A.~F. and Heafield, K.
\newblock Sparse communication for distributed gradient descent.
\newblock In \emph{Empirical Methods in Natural Language Processing (EMNLP)},
  2017.

\bibitem[Alistarh et~al.(2017)Alistarh, Grubic, Li, Tomioka, and
  Vojnovic]{alistarh2017qsgd}
Alistarh, D., Grubic, D., Li, J., Tomioka, R., and Vojnovic, M.
\newblock {QSGD}: Communication-efficient {SGD} via gradient quantization and
  encoding.
\newblock In \emph{Advances in Neural Information Processing Systems (NIPS)},
  pp.\  1707--1718. 2017.

\bibitem[Chang \& Lin(2011)Chang and Lin]{chang2011libsvm}
Chang, C.-C. and Lin, C.-J.
\newblock {LIBSVM}: A library for support vector machines.
\newblock \emph{ACM Transactions on Intelligent Systems and Technology (TIST)},
  2\penalty0 (27):\penalty0 1--27, 2011.

\bibitem[Chaturapruek et~al.(2015)Chaturapruek, Duchi, and
  R\'{e}]{chaturapruek2015asynchronous}
Chaturapruek, S., Duchi, J.~C., and R\'{e}, C.
\newblock Asynchronous stochastic convex optimization: the noise is in the
  noise and sgd don't care.
\newblock In \emph{Advances in Neural Information Processing Systems (NIPS)},
  pp.\  1531--1539. 2015.

\bibitem[De~Sa et~al.(2017)De~Sa, Feldman, R{\'e}, and
  Olukotun]{desa2017understanding}
De~Sa, C., Feldman, M., R{\'e}, C., and Olukotun, K.
\newblock Understanding and optimizing asynchronous low-precision stochastic
  gradient descent.
\newblock In \emph{Annual International Symposium on Computer Architecture
  (ISCA)}, pp.\  561--574, 2017.

\bibitem[Dean et~al.(2012)Dean, Corrado, Monga, Chen, Devin, Mao, aurelio
  Ranzato, Senior, Tucker, Yang, Le, and Ng]{dean2012large}
Dean, J., Corrado, G., Monga, R., Chen, K., Devin, M., Mao, M., aurelio
  Ranzato, M., Senior, A., Tucker, P., Yang, K., Le, Q.~V., and Ng, A.~Y.
\newblock Large scale distributed deep networks.
\newblock In \emph{Advances in Neural Information Processing Systems (NIPS)},
  pp.\  1223--1231. 2012.

\bibitem[He et~al.(2016)He, Zhang, Ren, and Sun]{he2016deep}
He, K., Zhang, X., Ren, S., and Sun, J.
\newblock Deep residual learning for image recognition.
\newblock In \emph{IEEE Conference on Computer Vision and Pattern Recognition
  (CVPR)}, pp.\  770--778, June 2016.

\bibitem[Huffman(1952)]{huffman1952method}
Huffman, D.~A.
\newblock A method for the construction of minimum-redundancy codes.
\newblock \emph{Proceedings of the IRE}, 40\penalty0 (9):\penalty0 1098--1101,
  Sept 1952.

\bibitem[Krizhevsky(2009)]{krizhevsky2009learning}
Krizhevsky, A.
\newblock Learning multiple layers of features from tiny images.
\newblock Technical report, 2009.

\bibitem[Lin et~al.(2018)Lin, Han, Mao, Wang, and Dally]{lin2018deep}
Lin, Y., Han, S., Mao, H., Wang, Y., and Dally, W.~J.
\newblock Deep gradient compression: Reducing the communication bandwidth for
  distributed training.
\newblock In \emph{International Conference on Learning Representations
  (ICLR)}, Apr 2018.

\bibitem[Recht et~al.(2011)Recht, Re, Wright, and Niu]{recht2011hogwild}
Recht, B., Re, C., Wright, S., and Niu, F.
\newblock Hogwild: A lock-free approach to parallelizing stochastic gradient
  descent.
\newblock In \emph{Advances in Neural Information Processing Systems (NIPS)},
  pp.\  693--701. 2011.

\bibitem[Russakovsky et~al.(2015)Russakovsky, Deng, Su, Krause, Satheesh, Ma,
  Huang, Karpathy, Khosla, Bernstein, Berg, and
  Fei-Fei]{russakovsky2015imagenet}
Russakovsky, O., Deng, J., Su, H., Krause, J., Satheesh, S., Ma, S., Huang, Z.,
  Karpathy, A., Khosla, A., Bernstein, M., Berg, A.~C., and Fei-Fei, L.
\newblock {I}mage{N}et large scale visual recognition challenge.
\newblock \emph{International Journal of Computer Vision (IJCV)}, 115\penalty0
  (3):\penalty0 211--252, 2015.

\bibitem[Seide et~al.(2014)Seide, Fu, Droppo, Li, and Yu]{seide2014bit}
Seide, F., Fu, H., Droppo, J., Li, G., and Yu, D.
\newblock 1-bit stochastic gradient descent and its application to
  data-parallel distributed training of speech {DNN}s.
\newblock In \emph{Interspeech}, 2014.

\bibitem[Strom(2015)]{strom2015calable}
Strom, N.
\newblock Scalable distributed {DNN} training using commodity {GPU} cloud
  computing.
\newblock In \emph{Interspeech}, 2015.

\bibitem[Wangni et~al.(2017)Wangni, Wang, Liu, and Zhang]{wangni2017gradient}
Wangni, J., Wang, J., Liu, J., and Zhang, T.
\newblock Gradient sparsification for communication-efficient distributed
  optimization.
\newblock \emph{CoRR}, abs/1710.09854, 2017.

\bibitem[Wen et~al.(2017)Wen, Xu, Yan, Wu, Wang, Chen, and Li]{wen2017terngrad}
Wen, W., Xu, C., Yan, F., Wu, C., Wang, Y., Chen, Y., and Li, H.
\newblock {T}ern{G}rad: Ternary gradients to reduce communication in
  distributed deep learning.
\newblock In \emph{Advances in Neural Information Processing Systems (NIPS)},
  pp.\  1508--1518. 2017.

\bibitem[Yan et~al.(2015)Yan, Ruwase, He, and Chilimbi]{yan2015performance}
Yan, F., Ruwase, O., He, Y., and Chilimbi, T.
\newblock Performance modeling and scalability optimization of distributed deep
  learning systems.
\newblock In \emph{ACM SIGKDD International Conference on Knowledge Discovery
  and Data Mining (KDD)}, pp.\  1355--1364, 2015.

\bibitem[Zhang et~al.(2017)Zhang, Li, Kara, Alistarh, Liu, and
  Zhang]{zhang2017zipml}
Zhang, H., Li, J., Kara, K., Alistarh, D., Liu, J., and Zhang, C.
\newblock {Z}ip{ML}: Training linear models with end-to-end low precision, and
  a little bit of deep learning.
\newblock In \emph{International Conference on Machine Learning (ICML)}, pp.\
  4035--4043, Aug 2017.

\bibitem[Zhao \& Li(2016)Zhao and Li]{zhao2016fast}
Zhao, S.-Y. and Li, W.-J.
\newblock Fast asynchronous parallel stochastic gradient descent: A lock-free
  approach with convergence guarantee.
\newblock In \emph{AAAI Conference on Artificial Intelligence (AAAI)}, pp.\
  2379--2385, 2016.

\bibitem[Zheng et~al.(2017)Zheng, Meng, Wang, Chen, Yu, Ma, and
  Liu]{zheng2017asynchronous}
Zheng, S., Meng, Q., Wang, T., Chen, W., Yu, N., Ma, Z.-M., and Liu, T.-Y.
\newblock Asynchronous stochastic gradient descent with delay compensation.
\newblock In \emph{International Conference on Machine Learning (ICML)}, pp.\
  4120--4129, Aug 2017.

\bibitem[Zhou et~al.(2016)Zhou, Ni, Zhou, Wen, Wu, and Zou]{zhou2016dorefa}
Zhou, S., Ni, Z., Zhou, X., Wen, H., Wu, Y., and Zou, Y.
\newblock {D}o{R}e{F}a-{N}et: Training low bitwidth convolutional neural
  networks with low bitwidth gradients.
\newblock \emph{CoRR}, abs/1606.06160, 2016.

\end{thebibliography}
\bibliographystyle{icml2018}

\clearpage
\onecolumn

\section{Proof for Lemma 2}

\begin{proof}
By re-organizing the update rule of accumulated quantization error, we have:
\begin{equation}
\begin{split}
\mathbf{h}_{p}^{( t )} &= \beta \mathbf{h}_{p}^{( t - 1 )} + ( \mathbf{g}_{p}^{( t - 1 )} - \tilde{\mathbf{g}}_{p}^{( t - 1 )} ) \\
&= \beta \mathbf{h}_{p}^{( t - 1 )} + ( - \alpha \mathbf{h}_{p}^{( t - 1 )} - \boldsymbol{\varepsilon}_{p}^{( t - 1 )} ) \\
&= ( \beta - \alpha ) \cdot \mathbf{h}_{p}^{( t - 1 )} - \boldsymbol{\varepsilon}_{p}^{( t - 1 )} \\
&= - \sum_{t' = 0}^{t - 1} ( \beta - \alpha )^{t - t' - 1} \cdot \boldsymbol{\varepsilon}_{p}^{( t' )}
\end{split}
\label{eqn:update_rule_h}
\end{equation}
which indicates that $\mathbf{h}_{p}^{( t )}$ is the linear combination of all the previous quantization errors.

Taking the expectation of squared $l_{2}$-norm of both sides of the second to the last equality in (\ref{eqn:update_rule_h}), we have:
\begin{equation}
\begin{split}
\mathbb{E} \| \mathbf{h}_{p}^{( t )} \|_{2}^{2} &= \mathbb{E} \| ( \beta - \alpha ) \cdot \mathbf{h}_{p}^{( t - 1 )} - \boldsymbol{\varepsilon}_{p}^{( t - 1 )} \|_{2}^{2} \\
&= ( \beta - \alpha)^{2} \cdot \mathbb{E} \| \mathbf{h}_{p}^{( t - 1 )} \|_{2}^{2} + \mathbb{E} \| \boldsymbol{\varepsilon}_{p}^{( t - 1 )} \|_{2}^{2}
\end{split}
\label{eqn:decomp_h_var_bound}
\end{equation}
and the last equality holds due to the independence between $\mathbf{h}_{p}^{( t - 1 )}$ and $\boldsymbol{\varepsilon}_{p}^{( t - 1 )}$ (recall that all quantization errors are \iid random noises).

Since the quantization error $\boldsymbol{\varepsilon}_{p}^{( t - 1 )}$ have the following variance bound (from Theorem 1):
\begin{equation}
\begin{split}
\mathbb{E} \| \boldsymbol{\varepsilon}_{p}^{( t - 1 )} \|_{2}^{2} &\le \gamma \cdot \mathbb{E} \| \mathbf{g}_{p}^{( t - 1 )} + \alpha \mathbf{h}_{p}^{( t - 1 )} \|_{2}^{2} \\
&= \gamma \cdot \mathbb{E} \| \mathbf{g}_{p}^{( t - 1 )} \|_{2}^{2} + \alpha^{2} \gamma \cdot \mathbb{E} \| \mathbf{h}_{p}^{( t - 1 )} \|_{2}^{2} \\
&\le \gamma B + \alpha^{2} \gamma \cdot \mathbb{E} \| \mathbf{h}_{p}^{( t - 1 )} \|_{2}^{2}
\end{split}
\label{eqn:decomp_e_var_bound}
\end{equation}
where the second equality is also derived from the independence between $\mathbf{g}_{p}^{( t - 1 )}$ and $\mathbf{h}_{p}^{( t - 1 )}$.

By substituting (\ref{eqn:decomp_e_var_bound}) into (\ref{eqn:decomp_h_var_bound}), we have:
\begin{equation}
\begin{split}
\mathbb{E} \| \mathbf{h}_{p}^{( t )} \|_{2}^{2} &\le [ \alpha^{2} \gamma + ( \beta - \alpha)^{2} ] \cdot \mathbb{E} \| \mathbf{h}_{p}^{( t - 1 )} \|_{2}^{2} + \gamma B \\
&\le \sum_{t' = 0}^{t - 1} [ \alpha^{2} \gamma + ( \beta - \alpha)^{2} ]^{t - t' - 1} \cdot \gamma B \\
&= \frac{1 - \lambda^{t}}{1 - \lambda} \cdot \gamma B
\end{split}
\label{eqn:final_h_var_bound}
\end{equation}
where $\lambda = \alpha^{2} \gamma + ( \beta - \alpha )^{2}$.

By substituting (\ref{eqn:final_h_var_bound}) into the variance bound of quantization error at the $t$-th iteration, we have:
\begin{equation}
\begin{split}
\mathbb{E} \| \boldsymbol{\varepsilon}_{p}^{( t )} \|_{2}^{2} &\le \gamma B + \alpha^{2} \gamma \cdot \mathbb{E} \| \mathbf{h}_{p}^{( t )} \|_{2}^{2} \\
&\le \left[ 1 + \alpha^{2} \gamma \cdot \frac{1 - \lambda^{t}}{1 - \lambda} \right] \cdot \gamma B
\end{split}
\end{equation}
which completes the proof.
\end{proof}

\section{Proof for Theorem 1}

\begin{proof}
Recall the update rule in ECQ-SGD:
\begin{equation}
\mathbf{w}^{( t + 1 )} = \mathbf{w}^{( t )} - \eta ( \mathbf{A} \mathbf{w}^{( t )} + \mathbf{b} + \boldsymbol{\xi}^{( t )} + \alpha \mathbf{h}^{( t )} + \boldsymbol{\varepsilon}^{( t )} )
\end{equation}

By applying the optimality of $\mathbf{w}^{*}$, and subtracting $\mathbf{w}^{*}$ from both sides of the above equality, we arrive at (note that we introduce $\mathbf{H} = \mathbf{I} - \eta \mathbf{A}$ for simplicity):
\begin{equation}
\begin{split}
\mathbf{w}^{( t + 1 )} - \mathbf{w}^{*} &= \mathbf{H} ( \mathbf{w}^{( t )} - \mathbf{w}^{*} ) - \eta ( \boldsymbol{\xi}^{( t )} + \alpha \mathbf{h}^{( t )} + \boldsymbol{\varepsilon}^{( t )} ) \\
&= \mathbf{H}^{t + 1} ( \mathbf{w}^{( 0 )} - \mathbf{w}^{*} ) - \boldsymbol{\Psi}^{( t )} - \boldsymbol{\Phi}^{( t )}
\end{split}
\end{equation}
where:
\begin{equation}
\begin{split}
\boldsymbol{\Psi}^{( t )} &= \eta \sum_{t' = 0}^{t} \mathbf{H}^{t - t'} \boldsymbol{\xi}^{( t' )} \\
\boldsymbol{\Phi}^{( t )} &= \eta \sum_{t' = 0}^{t} \mathbf{H}^{t - t'} ( \alpha \mathbf{h}^{( t' )} + \boldsymbol{\varepsilon}^{( t' )} )
\end{split}
\end{equation}

Since each accumulated quantization error $\mathbf{h}^{( t' )}$ is the linear combination of all previous quantization errors:
\begin{equation}
\mathbf{h}^{( t' )} = \sum_{t'' = 0}^{t' - 1} ( \beta - \alpha )^{t' - t'' - 1} \cdot \boldsymbol{\varepsilon}^{( t'' )}
\end{equation}
we can further simplify $\boldsymbol{\Phi}^{( t )}$ as:
\begin{equation}
\boldsymbol{\Phi}^{( t )} = \eta \sum_{t' = 0}^{t} \boldsymbol{\Theta}^{( t' )} \boldsymbol{\varepsilon}^{( t' )}
\end{equation}
where:
\begin{equation}
\boldsymbol{\Theta}^{( t' )} = \mathbf{H}^{t - t'} - \sum_{t'' = t' + 1}^{t} \alpha ( \beta - \alpha )^{t'' - t' - 1} \mathbf{H}^{t - t''}
\end{equation}

Due to the independence between all the random noises ($\{ \boldsymbol{\xi}^{( t' )} \}$ and $\{ \boldsymbol{\varepsilon}^{( t' )} \}$), the expectation of squared Euclidean distance between $\mathbf{w}^{( t + 1 )}$ and $\mathbf{w}^{*}$ is bounded by:
\begin{equation}
\begin{split}
\mathbb{E} \| \mathbf{w}^{( t + 1 )} - \mathbf{w}^{*} \|_{2}^{2} &= \mathbb{E} \| \mathbf{H}^{t + 1} ( \mathbf{w}^{( 0 )} - \mathbf{w}^{*} ) \|_{2}^{2} + \eta^{2} \sum_{t' = 0}^{t} \left[ \mathbb{E} \| \mathbf{H}^{t - t'} \boldsymbol{\xi}^{( t' )} \|_{2}^{2} + \mathbb{E} \| \boldsymbol{\Theta}^{( t' )} \boldsymbol{\varepsilon}^{( t' )} \|_{2}^{2} \right] \\
&\le R^{2} \| \mathbf{H}^{t + 1} \|_{2}^{2} + \eta^{2} \sigma^{2} \sum_{t' = 0}^{t} \| \mathbf{H}^{t'} \|_{2}^{2} + \eta^{2} \mathbb{E} \| \boldsymbol{\varepsilon}^{( t )} \|_{2}^{2} + \eta^{2} \sum_{t' = 0}^{t - 1} \| \boldsymbol{\Theta}^{( t' )} \|_{2}^{2} \cdot \mathbb{E} \| \boldsymbol{\varepsilon}^{( t' )} \|_{2}^{2}
\end{split}
\end{equation}
which completes the proof.
\end{proof}

\section{Proof for Lemma 3}

\begin{proof}
Since $\mathbf{A} \succeq a_{1} \mathbf{I}$, and the learning rate satisfies $\eta a_{1} < 1$, we have $\mathbf{I} - \eta \mathbf{A} \preceq ( 1 - \eta a_{1} ) \mathbf{I}$, which implies that $( \mathbf{I} - \eta \mathbf{A} )^{t''} \preceq ( 1 - \eta a_{1} )^{t''} \mathbf{I}$ holds for any positive integer $t''$. Therefore, we can derive the following inequality:
\begin{equation}
\begin{split}
( \mathbf{I} - \eta \mathbf{A} )^{t} &= ( \mathbf{I} - \eta \mathbf{A} )^{t - t'} ( \mathbf{I} - \eta \mathbf{A} )^{t'} \\
&\preceq ( 1 - \eta a_{1} )^{t - t'} ( \mathbf{I} - \eta \mathbf{A} )^{t'}
\end{split}
\end{equation}

By substituting the above inequality into the definition of $\boldsymbol{\Theta}^{( t' )}$ ($t' < t$), we arrive at:
\begin{equation}
\begin{split}
\boldsymbol{\Theta}^{( t' )} &\preceq \left[ 1 - \sum_{t'' = t' + 1}^{t} \frac{\alpha ( \beta - \alpha )^{t'' - t' - 1}}{( 1 - \eta a_{1} )^{t'' - t'}} \right] \cdot ( \mathbf{I} - \eta \mathbf{A} )^{t - t'} \\
&= \left[ 1 - \frac{\alpha}{\beta - \alpha} \sum_{t'' = 1}^{t - t'} \left( \frac{\beta - \alpha}{1 - \eta a_{1}} \right)^{t''} \right] \cdot ( \mathbf{I} - \eta \mathbf{A} )^{t - t'} \\
&= \left[ 1 - \frac{\alpha}{1 - \eta a_{1}} \frac{1 - \nu^{t - t'}}{1 - \nu} \right] \cdot ( \mathbf{I} - \eta \mathbf{A} )^{t - t'}
\end{split}
\end{equation}
where $\nu = ( \beta - \alpha) / ( 1 - \eta a_{1} )$.
\end{proof}

\section{Proof for Lemma 4}

\begin{proof}
Here we use $\Delta t = t - t'$ to denote the time gap. With $\beta = 1 - \eta a_{1}$ and $0 < \alpha < \beta$, we have $\nu = \frac{\beta - \alpha}{1 - \eta a_{1}} \in \left( 0, 1 \right)$, which leads to:
\begin{equation}
\lim_{\Delta t \rightarrow \infty} \nu^{\Delta t} = 0
\end{equation}

Recall that the upper bound of reduction ratio is given by:
\begin{equation}
\frac{\tau^{( t - \Delta t)}}{\tau_{QSGD}^{( t - \Delta t )}} < \left( 1 - \frac{\alpha}{1 - \eta a_{1}} \cdot \frac{1 - \nu^{\Delta t}}{1 - \nu} \right)^{2} \cdot \left( 1 + \frac{\alpha^{2} \gamma}{1 - \lambda} \right)
\end{equation}
and substituting $\beta = 1 - \eta a_{1}$ into it, we arrive at:
\begin{equation}
\lim_{\Delta t \rightarrow \infty} \frac{\tau^{( t - \Delta t)}}{\tau_{QSGD}^{( t - \Delta t )}} = \left( 1 - \frac{\alpha}{1 - \eta a_{1}} \cdot \frac{1}{1 - \frac{1 - \eta a_{1} - \alpha}{1 - \eta a_{1}}} \right)^{2} \cdot \left( 1 + \frac{\alpha^{2} \gamma}{1 - \lambda} \right) = 0
\end{equation}
which completes the proof.
\end{proof}

\end{document}